\documentclass[5p]{myelsarticle}

\usepackage{lineno,hyperref}
\hypersetup{colorlinks=true}
\modulolinenumbers[5]

\usepackage{wrapfig}
\usepackage{multirow}
\usepackage[table]{xcolor}
\usepackage{amsmath,amssymb,amsfonts}
\usepackage{color}
\usepackage{CJK}
\usepackage{adjustbox}
\usepackage[font=footnotesize,labelfont=bf]{caption}
\usepackage[section]{placeins}
\usepackage{float}

\journal{NeuroImage}

\usepackage[draft]{fixme}

\DeclareMathOperator*{\argmin}{arg\,min}

\begin{document}

\begin{frontmatter}

\title{Fast Predictive Simple Geodesic Regression}

\author[unc_cs]{Zhipeng Ding}
\author[usc,upenn]{Greg Fleishman}
\author[unc_cs]{Xiao Yang}
\author[usc]{Paul Thompson}
\author[aus]{Roland Kwitt}
\author[unc_cs,unc_bric]{Marc Niethammer}
\author[]{The Alzheimer's Disease Neuroimaging Initiative\corref{adni}}

%

\address[unc_cs]{Department of Computer Science, University of North Carolina at Chapel Hill, USA}
\address[unc_bric]{Biomedical Research Imaging Center, University of North Carolina at Chapel Hill, USA}
\address[usc]{Imaging Genetics Center, University of Southern California, USA}
\address[upenn]{Department of Radiology, University of Pennsylvania, USA}
\address[aus]{Department of Computer Science, University of Salzburg, Austria}

\cortext[adni]{Data used in preparation of this article were obtained from the Alzheimer's Disease Neuroimaging Initiative (ADNI) database (\url{adni.loni.usc.edu}). As such, the investigators within the ADNI contributed to the design and implementation of ADNI and/or provided data but did not participate in analysis or writing of this report. A complete listing of ADNI investigators can be found at: \url{http://adni.loni.usc.edu/wp-content/uploads/how_to_apply/ADNI_Acknowledgement_List.pdf}}

\begin{abstract}
\noindent
Deformable image registration and regression are important tasks in medical image analysis. However, they are computationally expensive, especially when analyzing large-scale datasets that contain thousands of images. Hence, cluster computing is typically used, making the approaches dependent on such computational infrastructure. Even larger computational resources are required as study sizes increase. This limits the use of deformable image registration and regression for clinical applications and as component algorithms for other image analysis approaches. We therefore propose using a fast predictive approach to perform image registrations. In particular, we employ these fast registration predictions to \emph{approximate} a simplified geodesic regression model to capture longitudinal brain changes. The resulting method is orders of magnitude faster than the standard optimization-based regression model and hence facilitates large-scale analysis on a single graphics processing unit (GPU). We evaluate our results on 3D brain magnetic resonance images (MRI) from the ADNI datasets.
\end{abstract}


\end{frontmatter}


\section{Introduction}
\noindent
Longitudinal image data provides us with a wealth of information to study aging processes, brain development and disease progression. Such studies, for example ADNI~\cite{jack2015magnetic} and the Rotterdam study~\cite{ikram2015rotterdam}, involve analyzing thousands of images. In fact, even larger studies will be available in the near future. For example, the UK Biobank~\cite{biobankWebsite} targets on the order of 100,000 images once completed. With the number of images increasing, large-scale image analysis typically resorts to using compute clusters for parallel processing.
While this is, in principle, a viable solution, increasingly larger compute clusters will become necessary for such studies. Alternatively, more efficient algorithms can reduce computational requirements, which then facilitates computations on individual computers or much smaller compute clusters, interactive (e.g., clinical) applications, efficient algorithm development, and use of these efficient algorithms as components in more sophisticated analysis approaches (which may use them as part of iterative processes).


\vskip1ex
Image registration is a key task in medical image analysis to study deformations between images. Building on image registration approaches, image regression models~\cite{ref:georegression,hong2012simple,hong2012metamorphic,ref:nikhil,fletcher2013geodesic,hong2014time,singh2014splines,hong2014geodesic,singh2015splines,hong2016parametric} have been developed to analyze deformation trends in longitudinal imaging studies. One such approach is geodesic regression (GR)~\cite{ref:georegression,ref:nikhil,fletcher2013geodesic} which (for images) build on the large displacement diffeomorphic metric mapping model (LDDMM)~\cite{ref:lddmm}. In general, GR generalizes linear regression to Riemannian manifolds. When applied to longitudinal image data, it can compactly express spatial image transformations over time. However, the solution to the underlying optimization problem is computationally expensive. Hence, a simplified, approximate, GR approach has been proposed~\cite{ref:hong} (SGR) to decouple the computation of the regression geodesic into pairwise image registrations. However, even such a simplified GR approach would require months of computation time on a single graphics processing unit (GPU) to process thousands of 3D image registrations for large-scale imaging studies such as ADNI~\cite{jack2015magnetic}. The primary reason computational bottleneck for SGR are the optimization required to compute pair-wise registrations.

\begin{figure*}[!t]
\centering
\includegraphics[width=\textwidth]{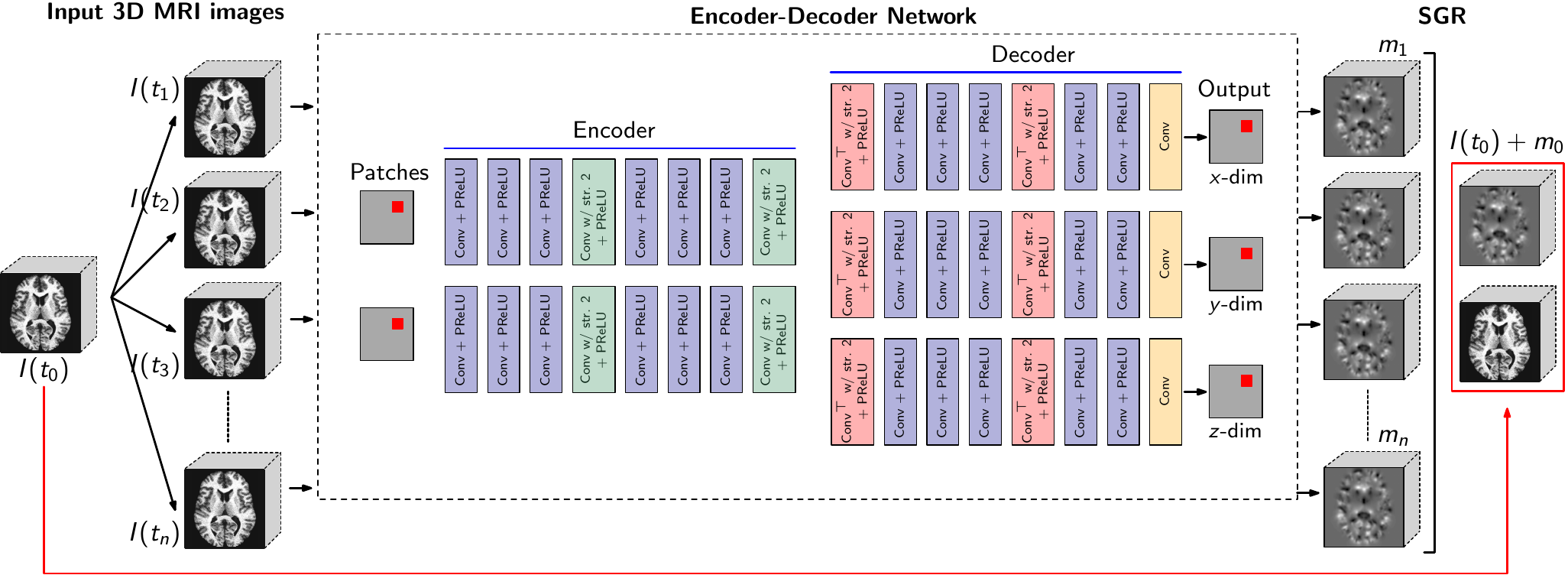}
\caption{Principle of fast predictive simple geodesic regression (FPSGR). In the encoder-decoder network (middle), the inputs are patches from the moving image and the target image at the \emph{same} spatial location; the outputs are the predicted initial momenta (i.e., $m_1,\ldots,m_n$) of the corresponding patches. \texttt{Conv}: Convolutional layer; \texttt{Conv}$^{T}$: transpose of convolutional layer. In the simple geodesic regression (SGR) part, all the pairwise initial momenta are \emph{averaged} according to Eq.~\eqref{eq:sgr} to produce the initial momentum of the regression geodesic (marked \textcolor{red}{red}).
\label{fig:1}}
\end{figure*}

\vskip1ex
Recently, efficient approaches have been proposed for deformable image registration~\cite{dinggang,miao2016real,nonrigid,ref:yang2016,quicksilver,zhang2017frequency}. In particular, for LDDMM, which is the basis of GR approaches for images, registrations can be dramatically sped up, by either working with finite-dimensional Lie algebras~\cite{zhang2015finite} and frequency diffeomorphisms~\cite{zhang2017frequency}, or by fast predictive image registration (FPIR)~\cite{ref:yang2016,quicksilver}. FPIR predicts the initial conditions (specifically, the initial momentum) of LDDMM, which fully characterize the geodesic and the spatial transformation using a \emph{learned} a patch-based deep regression model. Because numerical optimization of standard LDDMM registration is replaced by a {\it single} prediction step, followed by optional correction steps~\cite{quicksilver}, FPIR is dramatically faster than optimization-based LDDMM without compromising registration accuracy, as measured on several registration benchmarks~\cite{klein2009}.

\vskip1ex
Besides FPIR, other predictive image registration approaches have been proposed. Dosovitskiy et al.~\cite{flownet2015} use a convolutional neural network (CNN) to directly predict optical flow. Liu et al.~\cite{ref:liu} use an encoder-decoder network to synthesize video frames. Schuster et al.~\cite{schuster2016optical} investigate strategies to improve optical flow prediction via a CNN. Cao et al.~\cite{dinggang} use a sampling strategy and CNN regression to directly learn the mapping from moving and target image pairs to the final deformation field. Miao et al.~\cite{miao2016real} use CNN regression for 2D/3D rigid registration. Sokooti et al.\cite{nonrigid} use CNNs to directly predict a 3D displacement vector field from input image pairs. An unsupervised approach for image registration was proposed by de Vos et al.~\cite{dlmia2017_unsupervised}; here, the loss function is the image similarity measure between images themselves and a deformation is parameterized via a spatial transformer (which essentially amounts to a parameterized model of deformation in image registration) which generates the sought-for displacement vector field. In~\cite{hong2017fast}, Hong et al. employ a low-dimensional band-limited representation of velocity fields in Fourier space~\cite{zhang2015finite} to speed up SGR~\cite{ref:hong} for population-based image analysis.

\vskip1ex
In this work, we will build on FPIR, as it is a desirable approach for brain image registration for the following reasons: \emph{First}, FPIR predicts the initial momentum of LDDMM and therefore inherits the theoretical properties of LDDMM. Consequently, FPIR results in diffeomorphic transformations, even though predictions are computed in a patch-by-patch manner; this can not be guaranteed by most other prediction methods. \emph{Second}, patch-wise prediction allows for training of the prediction models based on a very small number of images, containing a large number of patches. \emph{Third}, by using a patch-wise approach, even high-resolution image volumes can be processed without running into memory issues on a GPU. \emph{Fourth},  none of the existing predictive methods address longitudinal data. However, as both FPIR and SGR are based on LDDMM, they naturally integrate and hence result in our proposed {\it fast predictive simple geodesic regression (FPSGR)} approach.

\vskip1ex
\noindent
Our \emph{contributions} can be summarized as follows:\\

\begin{description}
\item[Predictive geodesic regression] We use a fast predictive registration approach for image geodesic regression. 
Different to \cite{quicksilver}, we specifically validate that our approach can indeed capture the frequently subtle 
deformation trends of \emph{longitudinal} image data. 
\item[Large-scale dataset capability] Our predictive regression approach facilitates large-scale image regression within a short amount of time on a single GPU, instead of requiring months of computation time for standard optimization-based methods on a single computer, or on a compute cluster.
\item[Accuracy] We assess the accuracy of FPSGR by (1) studying linear models of atrophy scores (which are derived from the nonlinear SGR model) over time, as well as (2) correlations between atrophy scores and various diagnostic groups.
\item[Validation] We demonstrate the performance of FPSGR by analyzing $>6000$ images of the \texttt{ADNI-1} / \texttt{ADNI-2} 
datasets. For comparison, we also perform SGR using numerical optimization for the registrations, again 
on the complete \texttt{ADNI-1} / \texttt{ADNI-2} datasets.
\end{description}

%
%
%

\vskip1ex
\noindent
This work is an extension of a recent conference paper~\cite{ding2017}. In particular, all our experiments are now in 3D. We also added significantly more results to further explore the behavior of FPSGR in comparison to optimization-based SGR. In particular, we added (a) an example to visualize the performance of regression models and associated quantitative comparisons (Sec.~\ref{sec:regression_results}); (b) an analysis of local atrophy score correlated with clinical variables (Sec.~\ref{sec:atrophy}); (c) correlations within diagnostic groups (Sec.~\ref{sec:atrophy}); (d) a comparison with pairwise registration  (Sec.~\ref{sec:sgr_justification}); (e) and experiments on extrapolation on unseen data (Sec.~\ref{sec:forecast}, Sec.~\ref{sec:jacobian}).

\vskip1ex
\noindent
\textbf{Organization.} 
The remainder of this article is organized as follows: Sec.~\ref{sec:method} describes FPSGR, Sec.~\ref{sec:experimental_setup} discusses the experimental setup and the training of the prediction models. In Sec.~\ref{sec:discussion_of_experimental_results}, 
we present experimental results for 3D MR brain images. The paper concludes with a summary and an outlook on future work.

\section{Fast predictive simple geodesic regression}
\label{sec:method}
\noindent
Our fast predictive simple geodesic regression approach is a combination of two methods: \emph{First}, fast predictive image registration (FPIR) and, \emph{second}, integration of FPIR with simple geodesic regression (SGR). Both FPIR and SGR are based on the shooting formulation of LDDMM~\cite{ref:nikhil}; Fig.~\ref{fig:1} illustrates our overall approach. The individual components are described in the following.

\subsection{LDDMM}
\noindent
Shooting-based LDDMM and geodesic regression minimize
\begin{eqnarray}
\label{eq:gr}
E(I_0, m_0) = \frac{1}{2}\langle m_0, Km_0\rangle + \frac{1}{\sigma^2}\sum_i d^2(I(t_i), Y^i), \\ 
\notag
s.t. \quad m_t + \text{ad}^*_vm = 0, I_t + \nabla I^Tv = 0, m - Lv = 0,
\end{eqnarray}
where $I_0$ is the initial image (known for image-to-image registration and to be determined for geodesic regression), $m_0$ is the initial momentum, $K$ is a smoothing operator that connects velocity $v$ and momentum $m$ as $v = Km$ and $m = Lv$ with $K = L^{-1}$, $\sigma > 0$ is a weight, $Y^i$ is the measured image at time $t_i$ (there will be only one such image for image-to-image registration at $t=1$), and $d^2(I_1, I_2)$ denotes the image similarity measure between $I_1$ and $I_2$ (for example $L_2$ or geodesic distance); $\text{ad}^*$ is the dual of the negative Jacobi-Lie bracket of vector fields: $\text{ad}_vw = -[v, w] = Dvw - Dwv$ and $D$ denotes the Jacobian. The deformation of the  source image $I_0\circ\Phi^{-1}$ can be computed by solving $\Phi^{-1}_t + D\Phi^{-1}v=0,~\Phi^{-1}(0)=\text{id}$, where $\text{id}$ denotes the identity map.

\subsection{FPIR} 
\noindent 
Fast predictive image registration~\cite{ref:yang2016,quicksilver} aims at predicting the initial momentum, $m_0$, between a source and a target image patch-by-patch. Specifically, we use a deep encoder-decoder network to predict the patch-wise momentum. As shown in Fig. \ref{fig:1}, in 3D the inputs are two layers of $15 \times 15 \times 15$ image patches ($15 \times 15$ in 2D), where the two layers are from the source and target images respectively. Two patches are taken at the same position by two parallel encoders, which learn features independently. The output is the predicted initial momentum in the $x$, $y$ and $z$ directions (obtained by numerical optimization on the training samples). Basically, the network is split into an encoder and a decoder part. An \emph{encoder} consists of 2 blocks of three 3 $\times$ 3 $\times$ 3 convolutional layers with PReLU activations, followed by another 2 $\times$ 2 $\times$ 2 convolution+PReLU with a stride of two, serving as a ``pooling'' operation. The number of features in the first convolutional layer is 64 and increases to 128 in the second. In the \emph{decoder}, three parallel decoders share the same input generated from the encoder. Each decoder is the inverse of the encoder except for using 3D transposed convolution layers with a stride of two to perform ``unpooling'', and no non-linearity at the end. To speed up computations, we use patch pruning (i.e., for brain imaging, e.g., patches outside the brain are not predicted as the momentum is expected to be zero there) and a large pixel stride (e.g., 14 for $15 \times 15 \times 15$ patches) for the sliding window of the predicted patches.

\begin{figure*}[ht]
\centering
\includegraphics[width=\textwidth]{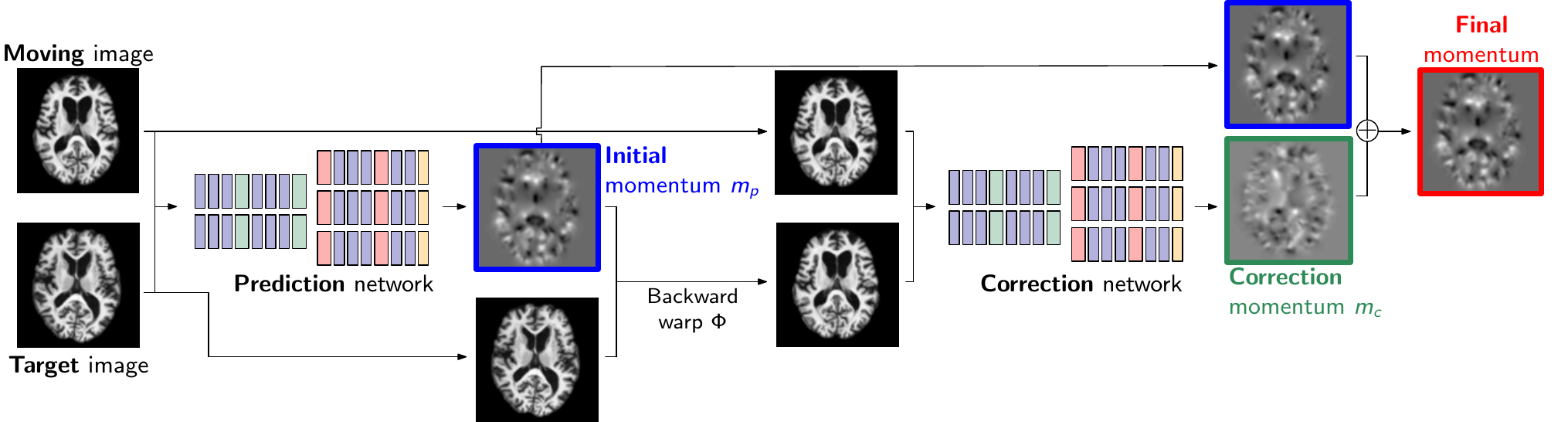}
\caption{Architecture of the prediction + correction network. Here, we use 2D images and the momentum in the $x$-direction for illustration. All images are 3D in our experiments.  (1) Predict the initial momentum $m_p$ and the corresponding backward deformation, $\Phi$; (2) Predict a correction of the initial momentum, $m_c$, based on the difference between the moving image and the warped-back target image. The final momentum is $m = m_p + m_c$. The correction network is trained based on the moving images and the warped-back target images of the training dataset.}
\label{fig:corr}
\end{figure*}

\subsection{Correction network}
\noindent 
We follow~\cite{quicksilver} and use a two-step approach to improve overall prediction accuracy. An additional correction step, i.e., a {\it correction network}, corrects the prediction of the initial prediction network. Fig.~\ref{fig:corr} illustrates this two-step approach graphically. The correction network has the same structure as the prediction network. Only the inputs and outputs differ. For the prediction network, the inputs are the original moving image and the original target image; output is the predicted initial momentum. For the correction network, the inputs are the original moving image and the warped target image; the output is the momentum difference.

\begin{table*}[!t] 
\centering
\begin{tabular}{ |c|c|c|c|c|c|c|c|c|}
\hline 
\multicolumn{8}{|c|}{\textbf{3D Longitudinal Test Case Deformation Error [pixel]}} \\ 
\hline 
\emph{Data Percentile} & 0.3\% & 5\% & 25\% & 50\% & 75\% & 95\% & 99.7\% \\ 
\hline 
Longitudinal Training& \cellcolor{green!30}\textbf{0.0156} & \cellcolor{green!30}\textbf{0.0407} & \cellcolor{green!30}\textbf{0.0761} & \cellcolor{green!30}\textbf{0.1098} & \cellcolor{green!30}\textbf{0.1559}  & \cellcolor{green!30}\textbf{0.2681}  & \cellcolor{green!30}\textbf{0.3238} \\ 
\hline
Cross-sectional Training& 0.0544 & 0.1424 & 0.2641 & 0.3723 & 0.5067  & 0.7502  & 0.8425 \\
\hline
\multicolumn{8}{|c|}{\textbf{3D Cross-sectional Test Case Deformation Error [pixel]}} \\ 
\hline 
\emph{Data Percentile} & 0.3\% & 5\% & 25\% & 50\% & 75\% & 95\% & 99.7\% \\ 
\hline 
Longitudinal Training& 0.1694 & 0.4802 & 1.0765 & 1.7649 & 2.7630  & 4.8060  & 5.6826 \\ 
\hline
Cross-sectional Training& \cellcolor{green!30}\textbf{0.1123} & \cellcolor{green!30}\textbf{0.3024} & \cellcolor{green!30}\textbf{0.5863} & \cellcolor{green!30}\textbf{0.8737} & \cellcolor{green!30}\textbf{1.2743}  & \cellcolor{green!30}\textbf{2.2659}  & \cellcolor{green!30}\textbf{2.7836} \\
\hline
\end{tabular} 
\caption{Deformation error of longitudinal and cross-sectional models tested on longitudinal and cross-sectional data. 2-norm deformation errors in pixels w.r.t. the ground truth deformation obtained by numerical optimization for LDDMM. A prediction model trained with longitudinal registration performs better for longitudinal registrations. Conversely, a model trained based on cross-sectional registration is preferred for cross-sectional registrations.}
\label{tab:1}
\end{table*}

\subsection{SGR}
\noindent
Determining the initial image, $I_0$, and the initial momentum, $m_0$, of Eq.~\eqref{eq:gr} is computationally costly. However, in simple geodesic regression, the initial image is fixed to the \textit{first} image of a subject's longitudinal image set (left-most part of Fig. \ref{fig:1}). Furthermore, the similarity measure $d(\cdot,\cdot)$ is chosen as the geodesic distance between images and {\it approximated} so that the geodesic regression problem can be solved by computing pair-wise image registrations with respect to the first image. Specifically, we define the quadratic distance $d^2$ between two images $A$ and $B$ as 
\begin{align}
& d^2(A, B) = \frac{1}{2} \int_0^1\lVert v^*\rVert^2_L dt, \\
\notag
& \text{where} \, v^* = \argmin_v \frac{1}{2} \int_0^1 \lVert v\rVert^2_L dt + \frac{1}{\sigma^2} \lVert Q(1) - B\rVert_2^2, \\
\notag
& \text{s.t.} \quad Q_t + \nabla Q^Tv = 0, \text{and}~Q(0) = A\enspace. 
\label{eq:geo_dist}
\end{align}
Assume we have an image $I(t_0)$ at time $t_0$ as well as two images $A(t_i)$ and $B(t_i)$. Further, assume that the spatial transformation $\Phi_A$ maps $A(t_i)$ to $I(t_0)$ and $\Phi_B$ maps $B(t_i)$ to $I_0$. Then $A(t_i)=I(t_0)\circ\Phi_A^{-1}$ and $B(t_i)=I(t_0)\circ\Phi_B^{-1}$. Furthermore, assume that $\Phi$ maps $A(t_i)$ to $B(t_i)$, i.e., $B(t_i) = A(t_i)\circ\Phi^{-1}$. Then $\Phi=\Phi_B\circ\Phi_A^{-1}$. Assuming that the geodesic between $I(t_0)$ and $A(t_i)$ is parameterized by the initial velocity $v^A$ and between $I(t_0)$ and $B(t_i)$ by the initial velocity $v^B$ and that we travel between $I(t_0)$ and $A(t_i)$ in time $t_i-t_0$ (and similarly for $B(t_i)$) we can rewrite the map between $A(t_i)$ and $B(t_i)$ based on the exponential map as
\begin{equation}
  \Phi = \mathrm{Exp_{Id}}((t_i-t_0)v^B)\circ  \mathrm{Exp_{Id}}(-(t_i-t_0)v^A),
\end{equation}
which can be approximated to first order as
\begin{equation}
  \Phi \approx  \mathrm{Exp_{Id}}((t_i-t_0)(v^B-v^A)).
\end{equation}
Hence, the squared geodesic distance between the two images can be approximated as
\begin{equation}
  d^2(A(t_i),B(t_i))\approx \frac{1}{2}(t_i-t_0)^2\langle K(m^B-m^A),m^B-m^A\rangle,
\end{equation}
where $v^A = K m^A$ and $v^B = K m^B$.
Hence, Eq.~\eqref{eq:gr} becomes
\begin{align}
& E(\overline{I}, \overline{m}) = \frac{1}{2}\langle \overline{m}, K\overline{m}\rangle \notag\\
& + \frac{1}{2\sigma^2}\sum_i (t_i-t_0)^2 \langle K(\overline{m}-m_i),\overline{m}-m_i \rangle,
\label{eq:approx}
\end{align}
where $\overline{m}$ is the sought-for initial momentum of the regression geodesic and $m_i$ are the initial momenta corresponding to the geodesic connecting $\overline{I}$ (the starting image of the geodesic) and the measurements $Y_i$ in time $t_i-t_0$. Differentiating Eq.~\eqref{eq:approx} w.r.t. $\overline{m}$ results in
\begin{align}
\nabla_{\overline{m}} E = K [\overline{m} + \frac{1}{\sigma^2} \sum_i (t_i - t_0)^2(\overline{m}-m_i)] \overset{!}{=} 0.
\end{align}
Thus,
\begin{equation}
\overline{m} = \frac{\sum_i (t_i - t_0)^2m_i}{\sigma^2 + \sum_i (t_i - t_0)^2}.\label{eq:gr_m0}
\end{equation}
In practice, $\sigma^2$ is very small and can thus be omitted. Furthermore, $m_i$ is obtained by either registering $\overline{I}$ to $Y^i$ in unit time or, as in our FPSGR approach, by predicting the momenta $m_i$ via FPIR, denoted as $\widetilde{m}_i$. As Equation~\ref{eq:gr_m0} was derived assuming that images are transformed into each other in time $t_i-t_0$ instead of unit time, the obtained unit-time predicted momenta $\widetilde{m}_i$ correspond in fact to the approximation $\widetilde{m}_i\approx (t_i-t_0)m_i$. Finally, 
we obtain the approximated optimal $\overline{m}$ of the energy functional in Eq.~\eqref{eq:gr}, for a fixed $\overline{I}=I_0$ as
\begin{eqnarray}
\qquad \qquad \overline{m} \approx \frac{\sum_i (t_i-t_0)\widetilde{m}_i}{\sum_i (t_i-t_0)^2}\label{eq:sgr}.
\end{eqnarray}

\section{Setup / Training}
\label{sec:experimental_setup}
\noindent
All our experiments use 3D images from the \texttt{ADNI} dataset\footnote{Data used in the preparation of this article were obtained from the Alzheimer's Disease Neuroimaging Initiative (ADNI) database (\url{adni.loni.usc.edu}). The ADNI was launched in 2003 as a public-private partnership, led by Principal Investigator Michael W. Weiner, MD. The primary goal of ADNI has been to test whether serial magnetic resonance imaging (MRI), positron emission tomography (PET), other biological markers, and clinical and neuropsychological assessment can be combined to measure the progression of mild cognitive impairment (MCI) and early Alzheimer's disease (AD).} which consists of 6471 3D MR brain images of size  $220 \times 220 \times 220$ voxels. In particular, \texttt{ADNI-1} contains 3479 images from 833 subjects and \texttt{ADNI-2} contains 2992 images from 823 subjects. Images belong to various types of diagnostic categories which we will discuss later. 

\vskip1ex
\noindent
We perform the following two types of studies:
\begin{description}
\item[Registration] We assess our hypothesis that training FPIR on longitudinal data for longitudinal registrations is preferred over training using cross-sectional data. Vice versa, training FPIR on cross-sectional data for cross-sectional registrations is preferred over training using longitudinal data. Comparisons are with respect to registration results obtained by numerical optimization (i.e., LDDMM).
\item[Regression] As for regression, we compare linear models fitted to atrophy scores over time, where scores are either
obtained from FPSGR or optimization-based SGR. Additionally, we study correlations between atrophy scores and 
diagnostic groups. Our hypothesis is that FPSGR is accurate enough to achieve comparable performance to
optimization-based SGR, at much lower computational cost, in both situations.
\end{description}

\subsection{Training of the prediction models}
\noindent 
We use a randomly selected set of 120 patients' MRI images from \texttt{ADNI} for training the prediction models and to test the performance of FPIR. We use all of the \texttt{ADNI} data for our regression experiments.\\

\noindent
\textbf{Training for registration.} We randomly selected 120 subjects from \texttt{ADNI-1} and registered their baseline images to their 24 month follow-up images. We used the first 100 subjects for training and the remaining 20 subjects for testing. For \emph{longitudinal training}, we registered the baseline image of a subject to the subject's 24-month image. For \emph{cross-sectional training}, we registered a subject's baseline image to another subject's 24-month image. To assess the performance of prediction models trained on these two types of paired data, we (1) perform the same type of registrations on the held-out 20 subjects
and (2) 
compare the 2-norm of the deformation error computed from the output of the prediction models with respect to the result obtained by numerical optimization of LDDMM\footnote{LDDMM results are generated using a vector momentum formulation: \url{https://bitbucket.org/scicompanat/vectormomentum}} (which serves as the ``ground-truth''). Table~\ref{tab:1} shows the results which confirm our hypothesis that training the prediction model with longitudinal registration cases is preferred for longitudinal registration over training with cross-sectional data. The deformation error is very small for longitudinal training / testing which provides strong evidence that the predictive method exhibits performance comparable to the (costly) 
optimization-based LDDMM. 
Another interpretation of these results is, that it is beneficial to train a prediction model with deformations that are to be \emph{expected}, i.e., relatively small deformations for longitudinal registrations and larger deformations for cross-sectional registrations. As we are interested in longitudinal registrations for the \texttt{ADNI} data, we only train our 3D models using longitudinal registrations in the following.

\vskip1ex
\noindent
\textbf{Training for regression.}
The \texttt{ADNI-1} dataset contains 228 normal controls, 257 subjects with mild cognitive impairment (MCI), 149 with late mild cognitive impairment (LMCI), as well as 199 subjects suffering from Alzheimer's disease (AD). We randomly picked roughly 1/6 of patients from each diagnostic category to form a set of 139 subjects for training in \texttt{ADNI-1}, i.e., 38 normal controls, 43 MCI, 25 LMCI, as well as 33 AD subjects; this results in 139 subjects overall. The baseline images of each subject were registered to \emph{all} the later time-points within the same subject. To maintain the diagnostic ratio, we picked 
(out of all registrations) 45 registrations from the normal group, 50 registrations from the MCI group, 30 registrations from the LMCI group, and 40 registrations from  the AD group, resulting in 165 longitudinal registration cases for training.
\vskip0.5ex
The same strategy was applied to \texttt{ADNI-2}. In detail, \texttt{ADNI-2} contains 200 normal controls, 111 subjects with significant memory complaint (SMC), 182 subjects with early mild cognitive impairment (EMCI), 175 with late mild cognitive impairment (LMCI), and 155 subjects with Alzheimer's disease (AD). We picked 150 subjects
and 140 longitudinal registrations, consisting of 35 registrations from the control group, 20 registrations from the SMC group, 30 registrations from the EMCI group, 30 registrations from the LMCI group, and 25 registrations from the AD group. Note that
there are fewer registrations than subjects (140 \emph{vs.} 150) in this setup, as our priority is to maintain the overall diagnostic ratio.

\vskip0.5ex
For both, \texttt{ADNI-1} and \texttt{ADNI-2}, the remaining 5/6 of the data is used for testing.
We trained four prediction models and their four corresponding correction models, leading to eight prediction models in total, listed in Table \ref{table:predictionmodels}. We also note that the training sets within \texttt{ADNI-1} and \texttt{ADNI-2}, resp., were not overlapping.

\begin{table}[h!]
\centering{
\begin{tabular}{r|l}
\hline
\texttt{ADNI-1 Pred-1}			& Model v1 (no corr.) \\
\texttt{ADNI-1 Pred+Corr-1}		& Model v1 +1x corr. step\\
\texttt{ADNI-1 Pred-2}			& Model v2 (no corr.)\\
\texttt{ADNI-1 Pred+Corr-2}		& Model v2 +1x corr. step\\
\hline
\texttt{ADNI-2 Pred-1}			& Model v1 (no corr.) \\
\texttt{ADNI-2 Pred+Corr-1}		& Model v1 +1x corr. step\\ 
\texttt{ADNI-2 Pred-2}			& Model v2 (no corr.)\\
\texttt{ADNI-2 Pred+Corr-2}		& Model v2 +1x corr. step\\
\hline
\end{tabular}}
\caption{\label{table:predictionmodels} Overview of the trained prediction models.}
\end{table}


\begin{table}[!t] 
\centering
\normalsize
\begin{adjustbox}{max width=0.5\textwidth}
\begin{tabular}{ |c|p{0.7cm}|p{0.7cm}|p{0.7cm}|p{0.7cm}|p{0.7cm}|p{0.7cm}|p{0.7cm}|}
\hline 
\multicolumn{7}{|c|}{\textbf{Distribution of prediction cases in \texttt{ADNI-1}}} \\ 
\hline 
\cellcolor{black!30}{\texttt{Pred-1}} & 6mo & 12mo & 18mo & 24mo & 36mo & 48mo\\ 
\hline 
NC&  182& 172& 8& 151& 128& 38\\ 
\hline
MCI$^{*}$& 274 & 221 & 165 & 122 & 80  & 11 \\
\hline
AD& 153& 173& 66& 163& 69& 20\\
\hline
\textbf{Total}& 609& 566& 239& 436& 277& 69\\  
\hline\hline
\cellcolor{black!30}{\texttt{Pred-2}} & 6mo & 12mo & 18mo & 24mo & 36mo & 48mo\\ 
\hline
NC& 182& 168& 9& 144& 119& 33\\
\hline
MCI$^{*}$& 272& 224& 169& 124& 70& 10\\
\hline
AD& 152& 168& 64& 160& 67& 22\\
\hline
\textbf{Total}& 606& 560& 242& 428& 256& 65\\
\hline 
\end{tabular} 
\end{adjustbox}
\caption{Distribution of \texttt{Pred/Corr-1} and \texttt{Pred/Corr-2} cases in 
\texttt{ADNI-1}. MCI$^{*}$ is the combination of the MCI and LMCI diagnostic groups. 
18 month only has one diagnostic group.}
\label{tab:dist1}
\end{table}

\begin{table}[!t] 
\centering
\normalsize
\begin{adjustbox}{max width=0.5\textwidth}
\begin{tabular}{ |c|p{0.9cm}|p{0.9cm}|p{0.9cm}|p{0.9cm}|p{0.9cm}|}
\hline 
\multicolumn{6}{|c|}{\textbf{Distribution of prediction cases in \texttt{ADNI-2}}} \\ 
\hline 
\cellcolor{black!30}{\texttt{Pred-1}} & 3mo & 6mo & 12mo & 24mo & 36mo \\ 
\hline 
NC$^*$& 173 & 141 & 153 & 119 & 3 \\
\hline
MCI$^*$& 256& 232& 207& 142& 4\\
\hline
AD& 93& 95& 105& 66& 1\\
\hline
\textbf{Total}& 522&468& 465& 327& 8\\
\hline\hline
\cellcolor{black!30}{\texttt{Pred-2}} & 3mo & 6mo & 12mo & 24mo & 36mo\\ 
\hline
NC$^*$& 172& 142& 159& 122& 3\\
\hline
MCI$^*$& 257& 230& 202& 149& 4\\
\hline
AD& 94& 98& 101& 52& 1\\
\hline
\textbf{Total} & 523& 470& 462& 323& 8\\
\hline 
\end{tabular}
\end{adjustbox} 
\caption{Distribution of \texttt{Pred/Corr-1} and \texttt{Pred/Corr-2} cases in \texttt{ADNI-2}. Normal$^*$ denotes the combination of the Normal and SMC diagnostic groups; MCI$^*$ denotes the combination of the EMCI and LMCI diagnostic groups. Only a small number of images is available for the 36 months time point.}
\label{tab:dist2}
\end{table}

\subsection{Parameter selection}
\noindent 
We use the regularization kernel 
$$K = L^{-1} = (-a \nabla^2 - b\nabla(\nabla \cdot) + c)^{-2}$$ 
with $[a, b, c]$ set to $[1, 0, 0.1]$. The parameter $\sigma$, from equation~\eqref{eq:gr}, is set to $0.1$. 
We train our network (using  \texttt{ADAM}~\cite{adam}) over 10 epochs with a learning rate of $0.0001$. 

\subsection{Efficiency}
\noindent  
Once trained, the prediction models allow fast computations of registrations. We use a TITAN X (Pascal) GPU and \texttt{PyTorch}\footnote{\url{http://pytorch.org}} for our implementation of FPIR. For the 3D \texttt{ADNI-1} dataset ($220 \times 220 \times 220$ MR images), FPSGR took about one day to predict 2646 pairwise registrations (i.e., 25 [s]/prediction) and to compute the regression result. Optimization-based LDDMM\footnote{Here, we used 300 fixed iterations for each registration. 300 iterations can guarantee almost all the results converge. Note that the optimization-based LDDMM also uses a GPU implementation.} would require $\approx$ 40 days of runtime. Runtime for FPIR on \texttt{ADNI-2} is identical to \texttt{ADNI-1} as the images have the same spatial dimension.

\vskip1ex
Compared to the-state-of-art fast geodesic regression model~\cite{hong2017fast}, FPSGR is also at least twice as fast. The model in~\cite{hong2017fast} achieves $\approx 16$ times speed-up compared with SGR~\cite{ref:hong} for the same setting (parallel computing with the same number of cores). In our case, we achieve a more than 40 times speed-up compared with SGR for the same setting (a single GPU).

\section{Experimental results for 3D ADNI data}
\label{sec:discussion_of_experimental_results}

\begin{figure}[!t]
\begin{center}
\includegraphics[width=0.5\textwidth]{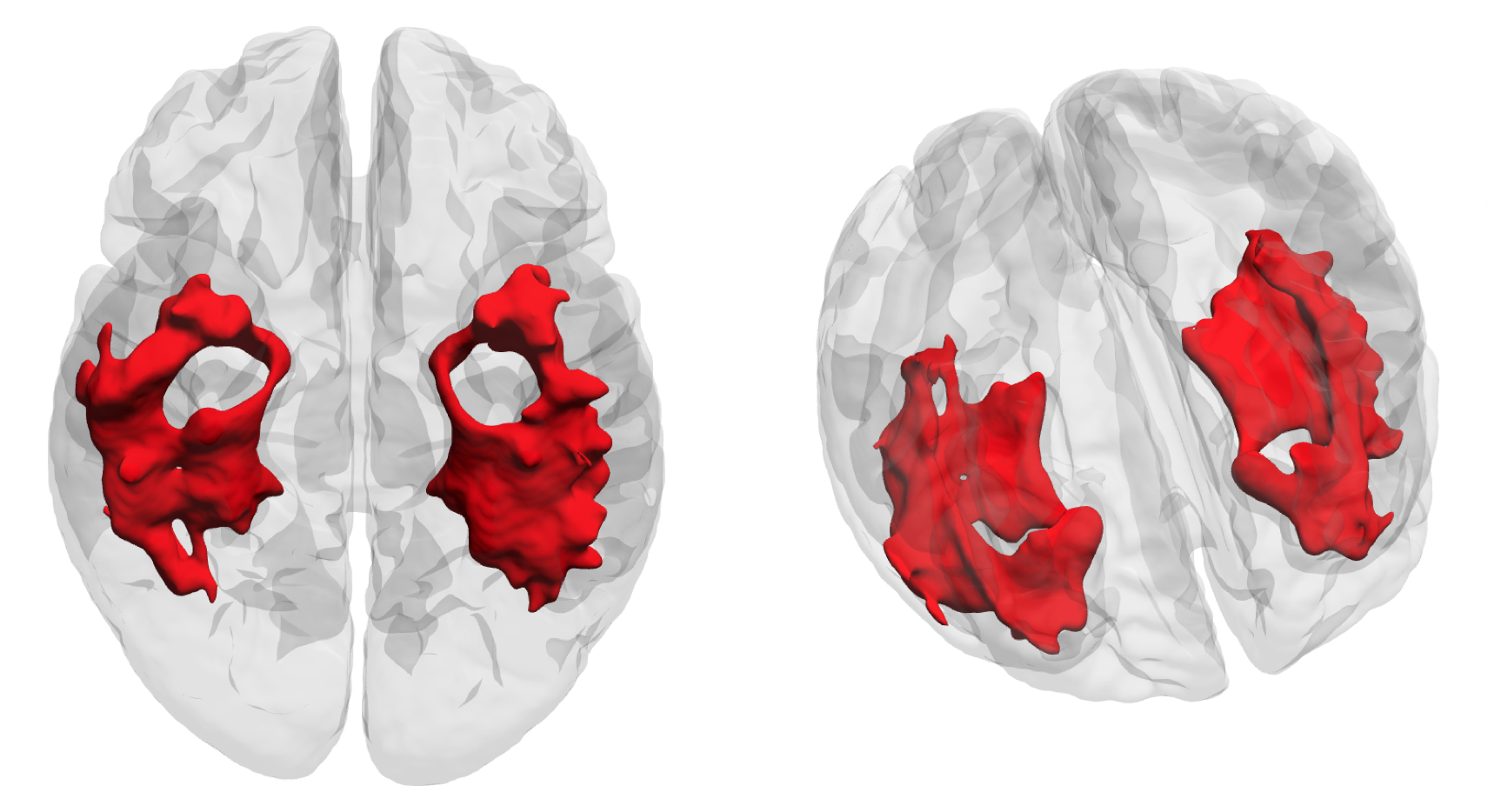}
\caption{Region of Interest (ROI) significantly associated with atrophy in AD used to compute atrophy scores.}
\label{fig:ROI}
\end{center}
\end{figure}

\noindent
For our experiments, we created 10 different (dataset, registration approach) combinations, each combination 
specifically designed to assess certain properties of our proposed strategy. These combinations
are described next.
\begin{itemize}
\item[1)] All subjects from the \texttt{ADNI-1} dataset in combination with optimization-based LDDMM.
\item[2)] Two subgroups of \texttt{ADNI-1} (i.e., different training data portions) in combination with FPSGR \emph{without} a correction network.
\item[3)] The same two subgroups as in 2), but in combination with FPSGR \emph{with} 
a correction network.
\item[4)] The same five groups of 1-3, but for \texttt{ADNI-2}.
\end{itemize}

Our general hypothesis is that the prediction models (for \texttt{ADNI-1/2}) show similar performance to optimization-based LDDMM and that using the correction network for the predictions improves results. To assess differences, we compare differences in deformations. Specifically, for every deformation produced by the different approaches, we compute its Jacobian determinant (JD). The JDs are then warped to a common coordinate system for the entire \texttt{ADNI} dataset using existing non-linear deformations from \cite{GMF_ISBI_matching,GMF_ISBI_optimization}. Each such spatially normalized JD is then averaged within a region where the rate of atrophy is significantly associated with Alzheimer's disease (AD), i.e., within a  
\emph{statistical region of interest (stat-ROI)} (see Fig.~\ref{fig:ROI}). Specifically, we quantify atrophy as 
\begin{equation}
  \left(1 - \frac{1}{|\omega|}  \int_{\omega} \text{det}(D \phi (x))~dx\right) \times 100
\end{equation} 
where $\text{det}(\cdot)$ denotes the determinant and $|\cdot|$ the cardinality/size of a set; $\omega$ is an area in the temporal lobes which was determined in prior studies~\cite{GMF_ISBI_matching,GMF_ISBI_optimization} to be significantly associated with accelerated atrophy in Alzheimer's disease. The resulting scalar value is an estimate of the relative volume change experienced by that region between the baseline and a follow-up image. Hence, its sign is positive when the region has lost volume over time and is negative if the region has gained volume over time. 

We limited our experiments to the applications in~\cite{Xue_ADNI1,Xue_ADNI2}, wherein nonlinear registration/regression is used to quantify atrophy within regions known to be associated to varying degrees with AD ($2$), mild cognitive impairment (MCI) ($1$) (including LMCI\footnote{We combine MCI and LMCI mainly because (a) the diagnostic changes available on the IDA website (\url{https://ida.loni.usc.edu/login.jsp}) only provide these three diagnostic groups; (b) to be consistent with the experiments conducted by Hua et al.~\cite{Xue_ADNI1}, where only Normal, MCI and AD were used as labels to classify \texttt{ADNI-1}. Hereafter, in all discussions of \texttt{ADNI-1}, MCI is a combination of MCI and LMCI of \texttt{ADNI-1}}), and normal ageing (NC: normal control) ($0$) in an elderly population. These are the diagnostic groups for \texttt{ADNI-1}. For \texttt{ADNI-2}, there are also 3 diagnostic categories\footnote{Similar to \texttt{ADNI-1}, a detailed diagnosis for \texttt{ADNI-2} is only available for the baseline images; MR images at later time points are only labeled as NC, MCI, and AD. Thus, we combine SMC and NC, as well as EMCI and LMCI to be consistent with the diagnostic changes in the \emph{ADNI Diagnosis Summary} available on the IDA website. Hereafter, in all discussions of \texttt{ADNI-2}, NC includes NC and SMC and MCI includes EMCI and LMCI.}: normal ageing ($0$) (including SMC), mild cognitive impairment (including EMCI and LMCI) ($1$), and AD ($2$).


\vskip1ex
\noindent
Specifically, we investigate the following \emph{six} questions:
\begin{itemize}
\item[\textbf{Q1})] Can the prediction models for regression qualitatively capture similar trends to the regression model obtained by numerical optimization? (Sec.~\ref{sec:regression_results})
\item[\textbf{Q2})]  Are atrophy measurements derived from FPSGR biased to overestimate or underestimate volume changes? (Sec.~\ref{sec:bias})
\item[\textbf{Q3})]  Are FPSGR atrophy measurements consistent with those derived from deformations via numerical optimization (LDDMM) which produced the training dataset? (Sec.~\ref{sec:atrophy})
\item[\textbf{Q4})]  Are regression results more stable and hence capture trends better than pairwise registrations? (Sec.~\ref{sec:sgr_justification})
\item[\textbf{Q5})]  Is the predictive power of the regression models strong enough to forecast deformations
for unseen future timepoints (Sec.~\ref{sec:forecast})
\item[\textbf{Q6})]  Do the prediction results capture expected trends in deformation? (Sec.~\ref{sec:jacobian})
\end{itemize}

If these experiments resolve favorably, then the substantially improved computational efficiency of FPSGR justifies its use for large-scale imaging studies. Tables~\ref{tab:dist1} and \ref{tab:dist2} show the distributions of the prediction cases per time-point and the diagnostic groups in \texttt{ADNI-1} and \texttt{ADNI-2}, respectively.

\subsection{Regression results}
\label{sec:regression_results}

\begin{figure*}[!t]
\centering
\includegraphics[width=\textwidth]{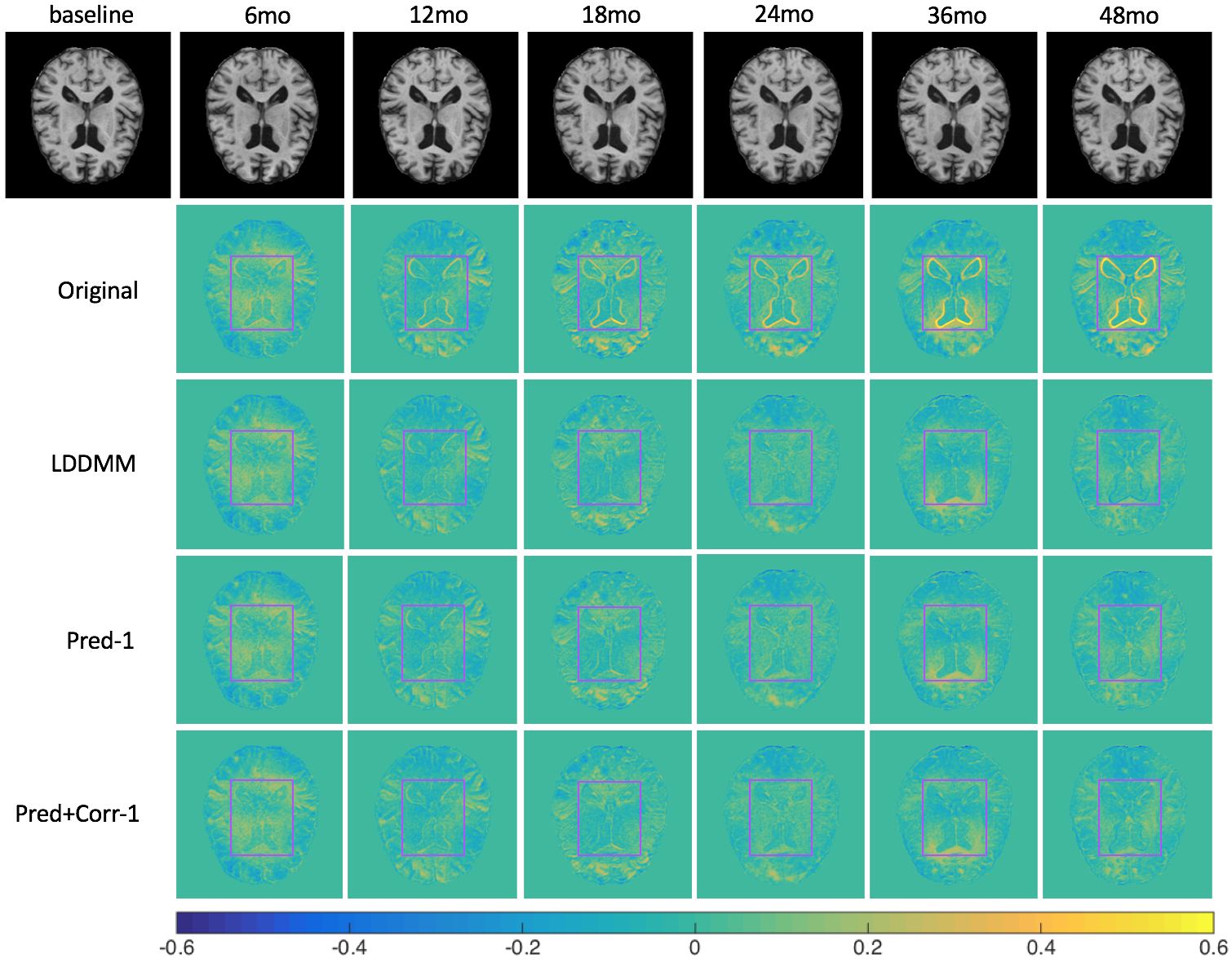}
\caption{Exemplary regression result: one subject with 6 follow-up images from the \texttt{ADNI-1} dataset. Image intensity range is [0, 2.49]. {\bf Top row:} Axial slices extracted from the 3D MR images at the same axial location for different months. {\bf Original:} intensity differences between the baseline image and its 6-month, 12-month, etc. follow-up image. {\bf LDDMM:} intensity differences between the acquired images in the top row and optimization-based regression results at each follow-up month(s). {\bf Pred-1:} intensity differences between the acquired images in the top row and the Pred-1 regression results at each follow-up month(s). {\bf Pred+Corr-1:} Same as for Pred-1, but using the Pred+Corr-1 regression model. Rectangles mark areas of major structural changes. Intensity differences are dramatically reduced, e.g., around the ventricles, demonstrating that these structural changes are captured by all three methods. The \textit{prediction models} (Pred-1, Pred+Corr-1) give very similar results to the regression results obtained by numerical optimization (LDDMM).}
\label{fig:example}
\end{figure*}

Table~\ref{tab:1} indicates that FPIR can predict deformation fields similar to the ones obtained using optimization-based LDDMM, even for the subtle changes seen in longitudinal imaging data. However, it remains to be seen how a predictive model performs for image regression. Fig.~\ref{fig:example} shows an exemplary regression result. In this specific case, large changes can be observed around the ventricles. To illustrate differences between the methods, Fig.~\ref{fig:example} shows regression results based on optimization-based LDDMM, for FPSGR \emph{without} a correction network, and for FPSGR \emph{with} a correction network. All three methods successfully capture the expanding ventricles and generally capture the image changes. Both FPSGR methods show results that are highly similar to SGR using optimization-based LDDMM. Hence, FPSGR is useful for longitudinal image regression. To further quantify the regression accuracy, we compute the overlay error between measured images and the images on the geodesic as
\begin{eqnarray}
\label{eq:overlay}
E_{overlay} (I_0 \circ \Phi^{-1}_{t_i}, Y_{i}) = \frac{1}{|\Omega|} \lVert I_0 \circ \Phi^{-1}_{t_i} - Y_{i}\rVert_{L_1}
\end{eqnarray}
where $\Omega$ is the brain area, $I_0 \circ \Phi^{-1}_{t_i}$ is the regressed image at time $t_i$ 
and $Y_{i}$ is the measured image at time $t_i$. Table~\ref{tab:overlay} shows the overlay error for the population of 100 subjects which includes all diagnostic groups in \texttt{ADNI-1}. 
Both FPSGR methods obtain results comparable with optimization-based LDDMM. This justifies the use of the proposed methods. The correction network generally increases the prediction accuracy over using the prediction network only.

\begin{table*}[!t] 
\centering
\normalsize
\begin{adjustbox}{max width=\textwidth}
\begin{tabular}{ |c|c|c|c|c|c|c|c|}
\hline 
&\multicolumn{6}{c|}{$E_{overlay} (I_0 \circ \Phi^{-1}_{t_i}, Y_{i})$} \\ 
\hline 
\texttt{Measured Images} & $I_{6mo}$ & $I_{12mo}$ & $I_{18mo}$ & $I_{24mo}$ & $I_{36mo}$ & $I_{48mo}$\\ 
\hline 
Original&  0.0770 $\pm$ 0.0212& 0.0764 $\pm$ 0.0207& 0.0890 $\pm$ 0.0220&0.0810 $\pm$ 0.0223& 0.0899 $\pm$ 0.0341& 0.0940 $\pm$ 0.0415\\ 
\hline
LDDMM& 0.0750 $\pm$ 0.0194& 0.0686 $\pm$ 0.0176 & 0.0734 $\pm$ 0.0190 & 0.0609 $\pm$ 0.0168 & 0.0628 $\pm$ 0.0177  & 0.0663 $\pm$ 0.0221 \\
\hline
Pred-1& 0.0754 $\pm$ 0.0213& 0.0694 $\pm$ 0.0182& 0.0742 $\pm$ 0.0195& 0.0621 $\pm$ 0.0188& 0.0654 $\pm$ 0.0184& 0.0698 $\pm$ 0.0238\\
\hline
Pred+Corr-1& 0.0754 $\pm$ 0.0211& 0.0691 $\pm$ 0.0182& 0.0734 $\pm$ 0.0192& 0.0615 $\pm$ 0.0166& 0.0642 $\pm$ 0.0188& 0.0688 $\pm$ 0.0235\\
\hline
\end{tabular} 
\end{adjustbox}
\caption{Mean+standard deviation of the overlay errors, see Eq.~\eqref{eq:overlay}, over 100 patients in \texttt{ADNI-1} dataset. Both prediction models exhibit performance comparable to optimization-based regression results (LDDMM). Including a correction network generally improves the performance of the 
prediction network.}
\label{tab:overlay}
\end{table*}

\subsection{Bias}
\label{sec:bias}

\begin{table}[!t]
\begin{center}
\begin{adjustbox}{max width=0.47\textwidth}
\scriptsize
\begin{tabular}{c|r|c|c|c}
\cellcolor{black!30}{\texttt{ADNI-1}}      &        & Slope                          & Intercept  &\#data \\
 \hline
 \multirow{ 6}{*}{NC-NC} &LDDMM-1 & [0.62, \textbf{0.70}, 0.78] & [-0.25,\textbf{-0.08}, 0.09] &\multirow{3}{*}{154}\\ 
 &Pred-1  & [0.37, \textbf{0.44}, 0.50] & [-0.21, \textbf{-0.08}, 0.05] \\  
 &Pred+Corr-1  & [0.61, \textbf{0.68}, 0.75] & [-0.15, \textbf{-0.01}, 0.13] \\
 \cline{2-5}
 &LDDMM-2 & [0.57, \textbf{0.66}, 0.75] & [-0.21, \textbf{-0.04}, 0.14] &\multirow{3}{*}{156}\\ 
 &Pred-2  & [0.43, \textbf{0.50}, 0.57] & [-0.16, \textbf{-0.02}, 0.11] \\  
 &Pred+Corr-2  & [0.51, \textbf{0.58}, 0.65] & [-0.12, \textbf{0.01}, 0.15] \\
 \hline
 \multirow{ 6}{*}{NC-MCI} &LDDMM-1 & [0.72, \textbf{0.94}, 1.16] & [-0.45, \textbf{-0.03}, 0.39] &\multirow{3}{*}{24}\\ 
 &Pred-1  & [0.39, \textbf{0.58}, 0.78] & [-0.43, \textbf{-0.05}, 0.33] \\  
 &Pred+Corr-1  & [0.71, \textbf{0.90}, 1.10] & [-0.40, \textbf{-0.01}, 0.37] \\
 \cline{2-5}
 &LDDMM-2 & [0.88, \textbf{1.19}, 1.50] & [-0.65, \textbf{-0.05}, 0.55] &\multirow{3}{*}{22}\\ 
 &Pred-2  & [0.72, \textbf{0.99}, 1.26] & [-0.68, \textbf{-0.16}, 0.36] \\  
 &Pred+Corr-2  & [0.80, \textbf{1.07}, 1.34] & [-0.66, \textbf{-0.14}, 0.38] \\
 \hline
 \multirow{ 6}{*}{MCI-MCI} &LDDMM-1 & [0.97, \textbf{1.17}, 1.38] & [-0.28, \textbf{0.05}, 0.39] &\multirow{3}{*}{146}\\ 
 &Pred-1  & [0.65, \textbf{0.80}, 0.96] & [-0.29, \textbf{-0.03}, 0.22] \\  
 &Pred+Corr-1  & [0.92, \textbf{1.09}, 1.26] & [-0.14, \textbf{0.14}, 0.42] \\
 \cline{2-5}
 &LDDMM-2 & [0.83, \textbf{1.00}, 1.17] & [-0.21, \textbf{0.06}, 0.33] &\multirow{3}{*}{148}\\
 &Pred-2  & [0.69, \textbf{0.82}, 0.96] & [-0.20, \textbf{0.02}, 0.24] \\  
 &Pred+Corr-2  & [0.77, \textbf{0.90}, 1.04] & [-0.15, \textbf{0.07}, 0.29] \\
 \hline
 \multirow{ 6}{*}{MCI-NC} &LDDMM-1 & [0.48, \textbf{0.72}, 0.96] & [-0.85, \textbf{-0.42}, 0.01] &\multirow{3}{*}{16}\\ 
 &Pred-1  & [0.26, \textbf{0.44}, 0.62] & [-0.61, \textbf{-0.29}, 0.03] \\  
 &Pred+Corr-1  & [0.51, \textbf{0.68}, 0.86] & [-0.52, \textbf{-0.20}, 0.13] \\
 \cline{2-5}
 &LDDMM-2 & [0.54, \textbf{0.79}, 1.03] & [-0.79, \textbf{-0.36}, 0.07] &\multirow{3}{*}{17}\\
 &Pred-2  & [0.40, \textbf{0.61}, 0.83] & [-0.62, \textbf{-0.24}, 0.14] \\  
 &Pred+Corr-2  & [0.49, \textbf{0.70}, 0.91] & [-0.59, \textbf{-0.21}, 0.17] \\
 \hline
\multirow{ 6}{*}{MCI-AD} &LDDMM-1 & [1.94, \textbf{2.10}, 2.27] & [-0.28, \textbf{0.02}, 0.31] &\multirow{3}{*}{148}\\ 
 &Pred-1  & [1.28, \textbf{1.40}, 1.53] & [-0.24, \textbf{-0.02}, 0.20] \\  
 &Pred+Corr-1  & [1.70, \textbf{1.84}, 1.98] & [-0.17, \textbf{0.08}, 0.33] \\
 \cline{2-5}
 &LDDMM-2 & [1.75, \textbf{1.92}, 2.09] & [-0.16, \textbf{0.14}, 0.44] &\multirow{3}{*}{147}\\
 &Pred-2  & [1.42, \textbf{1.56}, 1.70] & [-0.11, \textbf{0.14}, 0.39] \\  
 &Pred+Corr-2  & [1.49, \textbf{1.64}, 1.78] & [-0.08, \textbf{0.17}, 0.43] \\
 \hline
 \multirow{ 6}{*}{AD-AD} &LDDMM-1 & [1.97, \textbf{2.33}, 2.69] & [-0.17, \textbf{0.27}, 0.70] &\multirow{3}{*}{143}\\ 
 &Pred-1  & [1.23, \textbf{1.50}, 1.77] & [-0.13, \textbf{0.21}, 0.54] \\  
 &Pred+Corr-1  & [1.74, \textbf{2.05}, 2.35] & [-0.04, \textbf{0.33}, 0.70] \\
 \cline{2-5}
 &LDDMM-2 & [1.92, \textbf{2.28}, 2.65] & [-0.20, \textbf{0.24}, 0.68] &\multirow{3}{*}{140}\\
 &Pred-2  & [1.56, \textbf{1.85}, 2.15] & [-0.13, \textbf{0.22}, 0.57] \\  
 &Pred+Corr-2  & [1.65, \textbf{1.95}, 2.24] & [-0.10, \textbf{0.25}, 0.60] \\
 \hline
\cellcolor{black!30}{\texttt{ADNI-2}} & & Slope & Intercept\\
 \hline
\multirow{ 6}{*}{NC-NC} &LDDMM-1 & [0.55, \textbf{0.65}, 0.75] & [-0.08, \textbf{0.03}, 0.13] &\multirow{3}{*}{170}\\ 
 &Pred-1  & [0.41, \textbf{0.48}, 0.55] & [-0.03, \textbf{0.04}, 0.12] \\  
 &Pred+Corr-1  & [0.50, \textbf{0.57}, 0.65] & [-0.04, \textbf{0.05}, 0.13] \\
 \cline{2-5}
 &LDDMM-2 & [0.51, \textbf{0.62}, 0.72] & [-0.10, \textbf{0.01}, 0.12] & \multirow{3}{*}{175}\\ 
 &Pred-2  & [0.47, \textbf{0.55}, 0.62] & [-0.03, \textbf{0.05}, 0.13] \\  
 &Pred+Corr-2  & [0.35, \textbf{0.44}, 0.52] & [-0.09, \textbf{-0.00}, 0.08] \\
 \hline
\multirow{ 6}{*}{NC-MCI} &LDDMM-1 & [0.56, \textbf{0.79}, 1.02] & [-0.22, \textbf{0.01}, 0.25] & \multirow{3}{*}{16}\\ 
 &Pred-1  & [0.53, \textbf{0.68}, 0.82] & [-0.14, \textbf{0.01}, 0.16] \\  
 &Pred+Corr-1  & [0.63, \textbf{0.80}, 0.97] & [-0.16, \textbf{0.02}, 0.19] \\
 \cline{2-5}
 &LDDMM-2 & [0.62, \textbf{0.90}, 1.18] & [-0.32, \textbf{-0.02}, 0.28] & \multirow{3}{*}{17}\\ 
 &Pred-2  & [0.58, \textbf{0.77}, 0.97] & [-0.19, \textbf{0.01}, 0.22] \\  
 &Pred+Corr-2  & [0.46, \textbf{0.68}, 0.91] & [-0.25, \textbf{-0.02}, 0.22] \\
 \hline
\multirow{ 6}{*}{MCI-MCI} &LDDMM-1 & [0.71, \textbf{0.83}, 0.94] & [-0.13, \textbf{-0.00}, 0.12] & \multirow{3}{*}{184}\\ 
 &Pred-1  & [0.53, \textbf{0.61}, 0.68] & [-0.06, \textbf{0.02}, 0.10] \\  
 &Pred+Corr-1  & [0.64, \textbf{0.73}, 0.82] & [-0.08, \textbf{0.02}, 0.11] \\
 \cline{2-5}
 &LDDMM-2 & [0.71, \textbf{0.82}, 0.92] & [-0.14, \textbf{-0.02}, 0.09] & \multirow{3}{*}{183}\\ 
 &Pred-2  & [0.58, \textbf{0.66}, 0.73] & [-0.05, \textbf{0.03}, 0.12] \\  
 &Pred+Corr-2  & [0.50, \textbf{0.59}, 0.67] & [-0.12, \textbf{-0.02}, 0.07] \\
 \hline
\multirow{ 6}{*}{MCI-NC} &LDDMM-1 & [0.03, \textbf{0.39}, 0.74] & [-0.38, \textbf{0.05}, 0.47] & \multirow{3}{*}{16}\\ 
 &Pred-1  & [0.05, \textbf{0.29}, 0.52] & [-0.24, \textbf{0.05}, 0.33] \\  
 &Pred+Corr-1  & [0.08, \textbf{0.36}, 0.64] & [-0.28, \textbf{0.05}, 0.38] \\
 \cline{2-5}
 &LDDMM-2 & [0.14, \textbf{0.40}, 0.67] & [-0.28, \textbf{0.04}, 0.35] & \multirow{3}{*}{21}\\ 
 &Pred-2  & [0.24, \textbf{0.42}, 0.61] & [-0.17, \textbf{0.05}, 0.28] \\  
 &Pred+Corr-2  & [0.05, \textbf{0.26}, 0.48] & [-0.22, \textbf{0.03}, 0.29] \\
 \hline
 \multirow{ 6}{*}{MCI-AD} &LDDMM-1 & [1.65, \textbf{1.95}, 2.25] & [-0.21, \textbf{0.13}, 0.47] & \multirow{3}{*}{70}\\ 
 &Pred-1  &[1.09, \textbf{1.27}, 1.46] & [-0.12, \textbf{0.09}, 0.30] \\  
 &Pred+Corr-1  & [1.39, \textbf{1.62}, 1.85] & [-0.15, \textbf{0.11}, 0.37] \\
 \cline{2-5}
 &LDDMM-2 & [1.59, \textbf{1.91}, 2.23] & [-0.16, \textbf{0.19}, 0.53] & \multirow{3}{*}{65}\\ 
 &Pred-2  & [1.15, \textbf{1.35}, 1.56] & [-0.09, \textbf{0.14}, 0.36] \\  
 &Pred+Corr-2  & [1.20, \textbf{1.45}, 1.69] & [-0.13, \textbf{0.14}, 0.41] \\
 \hline
\multirow{ 6}{*}{AD-AD} &LDDMM-1 & [2.49, \textbf{2.76}, 3.04] & [-0.15, \textbf{0.07}, 0.30] & \multirow{3}{*}{101}\\ 
 &Pred-1  & [1.74, \textbf{1.90}, 2.07] & [-0.09, \textbf{0.04}, 0.18] \\  
 &Pred+Corr-1  & [2.14, \textbf{2.34}, 2.54] & [-0.09, \textbf{0.08}, 0.24] \\
 \cline{2-5}
 &LDDMM-2 & [2.72, \textbf{2.99}, 3.27] & [-0.15, \textbf{0.07}, 0.29] & \multirow{3}{*}{103}\\ 
 &Pred-2  & [1.97, \textbf{2.14}, 2.31] & [-0.07, \textbf{0.07}, 0.21] \\  
 &Pred+Corr-2  & [2.16, \textbf{2.36}, 2.56] & [-0.15, \textbf{0.02}, 0.18] \\
\hline
\end{tabular}
\end{adjustbox}
\caption{Slope and intercept values for simple linear regression of volume change over time. Our notation 
for \textit{slope} and \textit{intercept} indicate [lower bound of 95\% C.I., \textbf{point estimate}, upper bound of 95\% C.I.]. The interval of intercept estimates all contain zero. The slope changes between the different diagnostic groups. The \#data column lists the number of data points analyzed.} 
\label{tab:slope_and_intercept}

\end{center}
\end{table}

\begin{figure*}[!p]
\centerline{\includegraphics[width=0.9\textwidth]{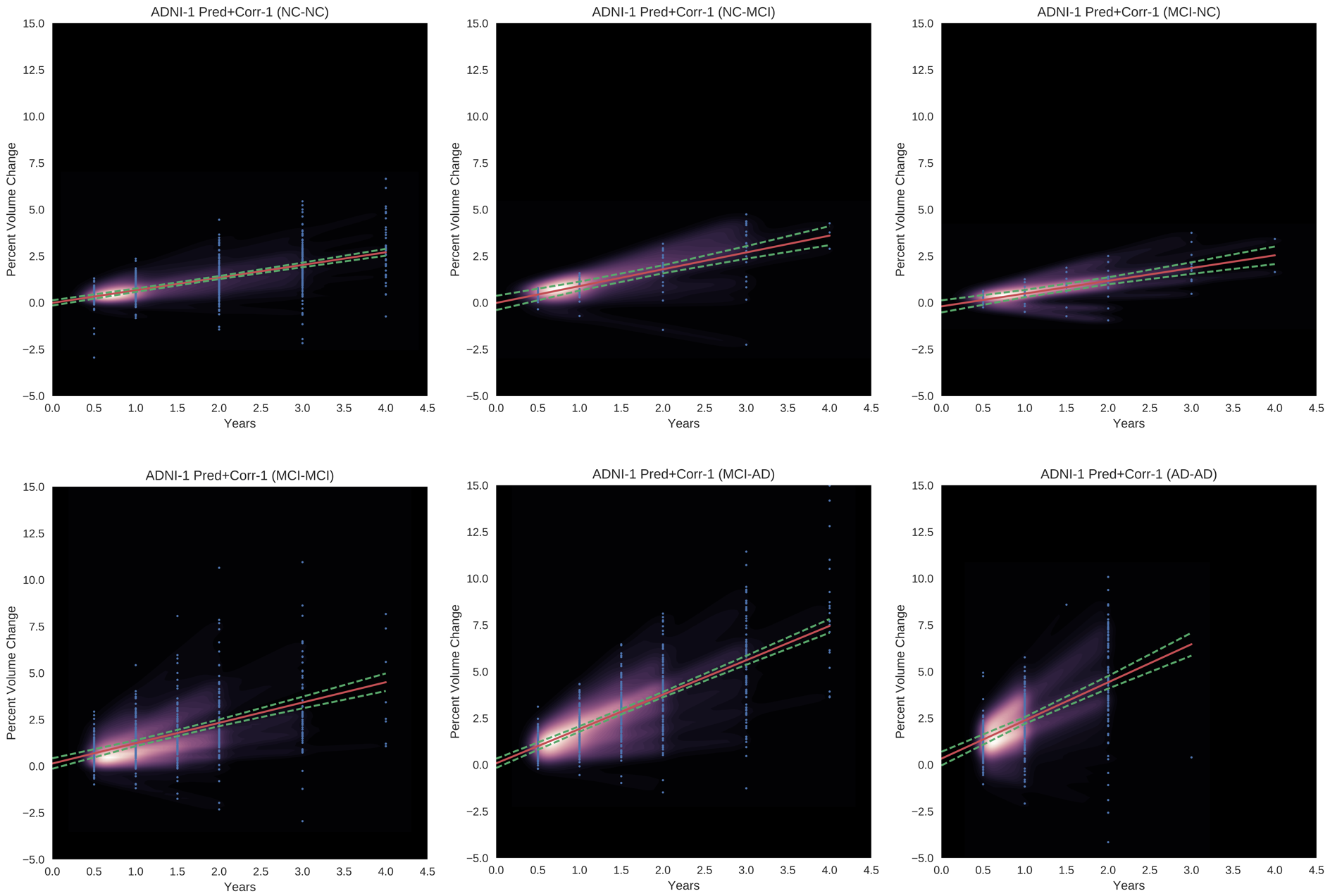}}
\centerline{\includegraphics[width=0.9\textwidth]{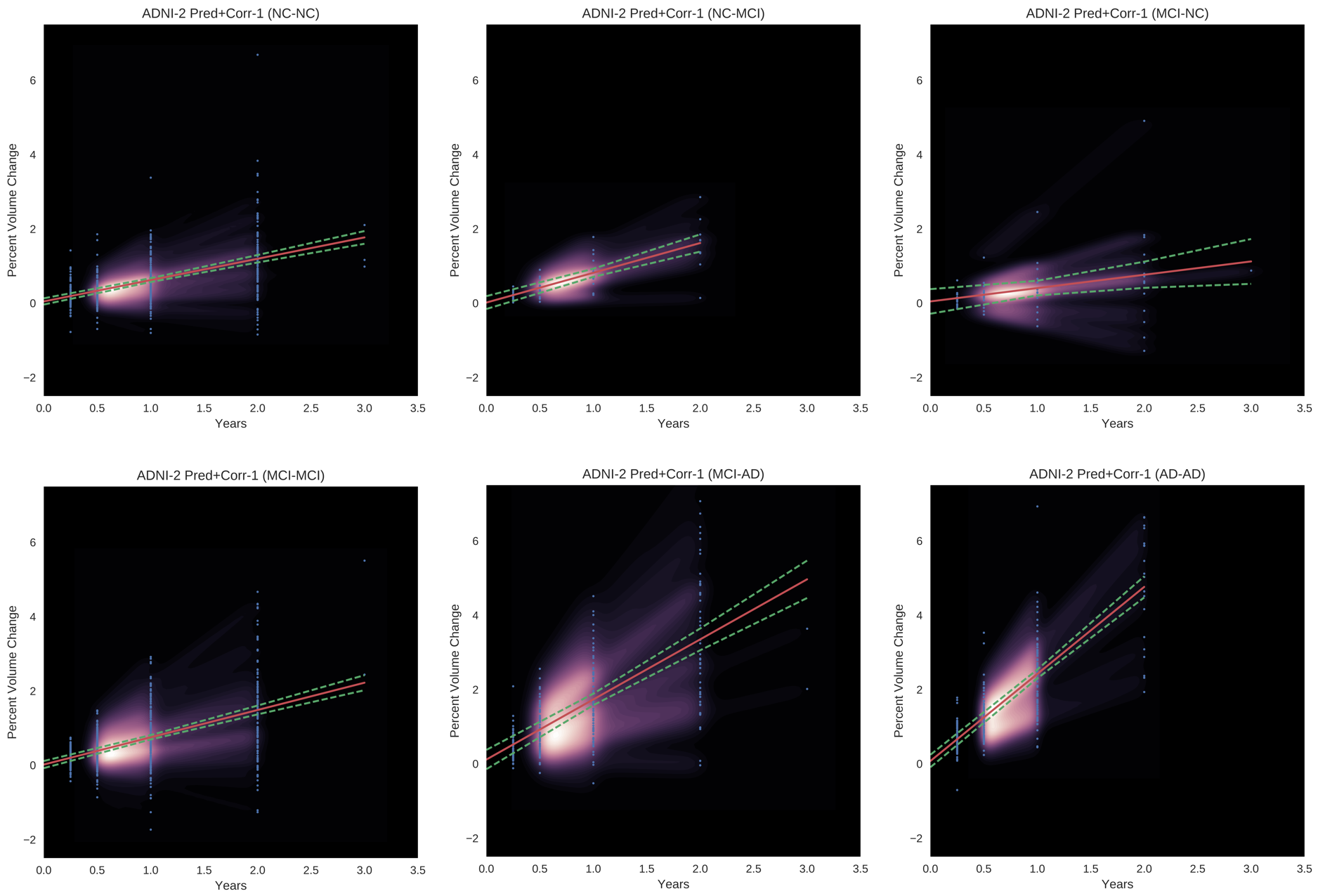}}
\caption{Linear regression of atrophy scores with respect to time for different diagnostic changes of \texttt{ADNI-1 Pred+Corr-1} and \texttt{ADNI-2 Pred+Corr-1}. Red line is the estimated regression line. green curves are the lower and upper bounds of the 95\% confidence interval. Blue dots indicate actual data points. Bright white / purple images indicate kernel density estimations for all real data points illustrating dominant longitudinal trends in the data.}
\label{fig:linear}
\end{figure*}

Estimates of atrophy are susceptible to bias \cite{Yushkevich_Bias}. To quantitatively assess this potential bias, we separately considered different diagnostic groups. Specifically, we considered six diagnostic change groups in our experiments: (1) NC for all time points (NC-NC), (2) starting with NC and changing to MCI or AD at a later time point (NC-MCI), (3) MCI for all time points (MCI-MCI), (4) starting with MCI and reversing to NC at later time points (MCI-NC), (5) starting with MCI and changing to AD at later time points (MCI-AD), and (6) AD for all the time points (AD-AD)\footnote{In \texttt{ADNI-1/ADNI-2}, there are two patients who show a reversion from AD to MCI. We omitted these cases in our experiment because the number of such cases is too small.}. In particular, we follow \cite{Xue_ADNI1} and fit a straight line (i.e., linear regression) through all atrophy measurements over time, conditioned on each diagnostic change category. The intercept term is an estimate of the atrophy one would measure when registering two scans acquired on the same day; hence it should be near zero and its 95\% confidence interval should contain zero. Quantitatively, Table~\ref{tab:slope_and_intercept} lists the slopes, intercepts, and 95\% confidence intervals for all ten groups of \texttt{ADNI-1} and \texttt{ADNI-2}, respectively. LDDMM-1 and LDDMM-2 denote the optimization-based results split into the same testing groups used for Pred-1 and Pred-2 to allow for a direct comparison. All of the results show intercepts that are near zero relative to the range of changes observed and all prediction intercept confidence intervals contain zero. For all diagnostic change groups the prediction and prediction+correction models exhibit more stable results than the optimization-based LDDMM method as indicated by the tighter confidence intervals. Furthermore, all slopes are positive, indicating average volume loss over time. This is consistent with expectations for an aging and neuro-degenerative population. The slopes capture increasing atrophy with disease severity. In \texttt{ADNI-1}/\texttt{ADNI-2}, we expect $\text{Slope}_{\text{NC-NC}} < \text{Slope}_{\text{MCI-NC}} < \text{Slope}_{\text{NC-MCI}} < \text{Slope}_{\text{MCI-AD}} < \text{Slope}_{\text{AD-AD}}$ and all six experimental groups (i.e. LDDMM-1, Pred-1, Pred+Corr-1, LDDMM-2, Pred-2, and Pred+Corr-2) are generally consistent with this expectation. Exceptions happen in \texttt{ADNI-2} for the NC-MCI and MCI-NC cases. As the number of subjects involved is relatively small, i.e., fewer than 20, compared with the other cases (roughly 100), one may speculate that this observation is caused by the limited number of data points for NC-MCI and MCI-NC as shown in the \#data column of Table~\ref{tab:slope_and_intercept}. However, the behavior within each starting diagnostic category, is consistent, i.e., for NC $\text{Slope}_{\text{NC-NC}} < \text{Slope}_{\text{NC-MCI}}$ and for MCI $\text{Slope}_{\text{MCI-NC}} < \text{Slope}_{\text{MCI-MCI}} < \text{Slope}_{\text{MCI-AD}}$. Hence, all six groups' slope results in \texttt{ADNI-1/ADNI-2} are generally consistent with our expectation (and also consistent with results in \cite{Xue_ADNI1}). The slope estimated from the prediction+correction results is larger than the slope estimated from the prediction model results and closer to the slope obtained from the optimization-based LDDMM results. This indicates that the correction network can improve prediction accuracy. Fig.~\ref{fig:linear} shows linear regression results for the estimated atrophy scores in \texttt{ADNI-1/2} for the Pred+Corr-1 model. Both the data points themselves (i.e., the atrophy scores), as well as kernel density estimates for the linear trends for each subject are shown. These results are consistent with the results of Table~\ref{tab:slope_and_intercept} discussed above. We conclude that (1) neither LDDMM optimization nor FPSGR produced deformations with significant bias to overestimate or underestimate volume change; (2) a linear model of atrophy scores generated by FPSGR can capture intrinsic volume change (i.e., slope) among different diagnostic change groups. Note that our LDDMM optimization results and the prediction results show the same trends.  Further, they are directly comparable as the results are based on the same test images (also for the atrophy measurements). 

\subsection{Atrophy}
\label{sec:atrophy}

\begin{table}[t!]
\begin{center}
\begin{adjustbox}{max width=0.46\textwidth} 
\scriptsize
\begin{tabular}{r|r|c|c|c|c|c}
 \cellcolor{black!30}{\texttt{ADNI-1}}      &        & MMSE        &$p$-value                  & DX  &$p$-value  &\#data\\
 \hline
 \multirow{ 6}{*}{6mo} &LDDMM-1 &-0.4957  &\cellcolor{purple!30}5.17e-39 &0.5140 &\cellcolor{purple!30}2.66e-42 & \multirow{3}{*}{608}\\ 
 &Pred-1  & -0.4642 & \cellcolor{purple!30}8.09e-34 & 0.4754 & \cellcolor{purple!30}1.30e-35  \\  
 &Pred+Corr-1 &\cellcolor{green!30}-0.5104  &\cellcolor{purple!30}1.22e-41  &\cellcolor{green!30}0.5259  &\cellcolor{purple!30}1.53e-44  \\
 \cline{2-7}
 &LDDMM-2 &-0.4667  &\cellcolor{purple!30}4.17e-34 &0.4814 &\cellcolor{purple!30}1.75e-36  &\multirow{3}{*}{606} \\ 
 &Pred-2  &-0.4711  &\cellcolor{purple!30}8.48e-35 &0.4849 &\cellcolor{purple!30}4.58e-37  \\  
 &Pred+Corr-2  &\cellcolor{green!30}-0.4734  &\cellcolor{purple!30}3.54e-35 &\cellcolor{green!30}0.4890 &\cellcolor{purple!30}9.67e-38\\
 \hline
 \multirow{ 6}{*}{12mo} &LDDMM-1 &-0.5749  &\cellcolor{purple!30}5.23e-51 &0.5313 &\cellcolor{purple!30}1.81e-42 &\multirow{3}{*}{565} \\ 
 &Pred-1  &-0.5328  &\cellcolor{purple!30}9.46e-43  &0.4898  &\cellcolor{purple!30}1.97e-35  \\  
 &Pred+Corr-1 &\cellcolor{green!30}-0.5799  &\cellcolor{purple!30}4.39e-52  &\cellcolor{green!30}0.5406  &\cellcolor{purple!30}3.44e-44  \\
 \cline{2-7}
 &LDDMM-2 &-0.5301  &\cellcolor{purple!30}6.81e-42 &0.5055 &\cellcolor{purple!30}1.17e-37  &\multirow{3}{*}{560} \\ 
 &Pred-2  &-0.5351  &\cellcolor{purple!30}9.79e-43 &0.5120 &\cellcolor{purple!30}1.11e-38  \\  
 &Pred+Corr-2  &\cellcolor{green!30}-0.5374  &\cellcolor{purple!30}3.73e-43 &\cellcolor{green!30}0.5155 &\cellcolor{purple!30}2.89e-39 \\
 \hline
 \multirow{ 6}{*}{18mo} &LDDMM-1 &\cellcolor{red!30}-0.4939  &\cellcolor{purple!30}4.86e-16 & \cellcolor{red!30}0.4776 & \cellcolor{purple!30}5.76e-15 &\multirow{3}{*}{238} \\ 
 &Pred-1  &-0.4659  &\cellcolor{purple!30}3.18e-14  & 0.4313 &\cellcolor{purple!30} 3.37e-12  \\  
 &Pred+Corr-1 &-0.4924  &\cellcolor{purple!30}6.16e-16  & 0.4643  & \cellcolor{purple!30}3.98e-14  \\
 \cline{2-7}
 &LDDMM-2 &-0.4385  &\cellcolor{purple!30}9.50e-13 & \cellcolor{red!30}0.4000 & \cellcolor{purple!30}1.12e-10 &\multirow{3}{*}{241} \\ 
 &Pred-2  &\cellcolor{yellow!30}-0.4389  &\cellcolor{purple!30}9.06e-13 & 0.3818 & \cellcolor{purple!30}8.80e-10 \\  
 &Pred+Corr-2  &-0.4384  &\cellcolor{purple!30}9.75e-13 & 0.3790 & \cellcolor{purple!30}1.19e-9  \\
 \hline
 \multirow{ 6}{*}{24mo} &LDDMM-1 &\cellcolor{red!30}-0.6064  &\cellcolor{purple!30}5.01e-45 &\cellcolor{red!30}0.5978 &\cellcolor{purple!30}1.69e-43 &\multirow{3}{*}{435} \\ 
 &Pred-1  &-0.5664  &\cellcolor{purple!30}2.83e-38  &0.5607  &\cellcolor{purple!30}2.18e-37  \\  
 &Pred+Corr-1 &-0.6001  &\cellcolor{purple!30}6.55e-44  &0.5943  &\cellcolor{purple!30}6.82e-43  \\
 \cline{2-7}
 &LDDMM-2 &-0.5822  &\cellcolor{purple!30}4.11e-40 &0.5534 &\cellcolor{purple!30}1.24e-35  &\multirow{3}{*}{427} \\ 
 &Pred-2  &\cellcolor{yellow!30}-0.5911  &\cellcolor{purple!30}1.41e-41 &\cellcolor{yellow!30}0.5714 &\cellcolor{purple!30}2.26e-38  \\  
 &Pred+Corr-2  &-0.5898  &\cellcolor{purple!30}2.28e-41 &0.5709 &\cellcolor{purple!30}2.65e-38 \\
 \hline
 \multirow{ 6}{*}{36mo} &LDDMM-1 &\cellcolor{red!30}-0.5142  &\cellcolor{purple!30}4.29e-20 &\cellcolor{red!30}0.5300 &\cellcolor{purple!30}1.81e-21 &\multirow{3}{*}{277} \\ 
 &Pred-1  &-0.4731  &\cellcolor{purple!30}7.38e-17  &0.4926  &\cellcolor{purple!30}2.42e-18  \\  
 &Pred+Corr-1 &-0.5069  &\cellcolor{purple!30}1.71e-19  &0.5296  &\cellcolor{purple!30}1.99e-21  \\
 \cline{2-7}
 &LDDMM-2 &-0.4334  &\cellcolor{purple!30}3.79e-13 &0.4815 &\cellcolor{purple!30}2.93e-16  &\multirow{3}{*}{256} \\ 
 &Pred-2  &\cellcolor{yellow!30}-0.4425  &\cellcolor{purple!30}1.07e-13 &\cellcolor{yellow!30}0.4894 &\cellcolor{purple!30}7.99e-17  \\  
 &Pred+Corr-2  &-0.4393  &\cellcolor{purple!30}1.67e-13 &0.4863 &\cellcolor{purple!30}1.34e-16  \\
 \hline
 \multirow{ 6}{*}{48mo} &LDDMM-1 &\cellcolor{red!30}-0.7456  &\cellcolor{purple!30}2.01e-13 &\cellcolor{red!30}0.6635 &\cellcolor{purple!30}5.20e-10 &\multirow{3}{*}{69} \\ 
 &Pred-1  &-0.7294  &\cellcolor{purple!30}1.18e-12  &0.6458  &\cellcolor{purple!30}2.08e-9  \\  
 &Pred+Corr-1 &-0.7443  &\cellcolor{purple!30}2.30e-13  &0.6575  &\cellcolor{purple!30}8.43e-10  \\
 \cline{2-7}
 &LDDMM-2 &-0.6889  &\cellcolor{purple!30}2.25e-10 &0.5927 &\cellcolor{purple!30}1.98e-7  & \multirow{3}{*}{65} \\ 
 &Pred-2  &-0.6995  &\cellcolor{purple!30}9.08e-11 &0.6048 &\cellcolor{purple!30}9.49e-8  \\  
 &Pred+Corr-2  &\cellcolor{green!30}-0.7005  &\cellcolor{purple!30}8.31e-11 &\cellcolor{green!30}0.6067 &\cellcolor{purple!30}8.49e-8 \\
 \hline
 \cellcolor{black!30}{\texttt{ADNI-2}}       &        & MMSE        &$p$-value                  & DX  &$p$-value  &\#data\\
 \hline
 \multirow{ 6}{*}{3mo} &LDDMM-1 &N/A  &N/A &0.4254 &\cellcolor{purple!30}2.34e-24 &\multirow{3}{*}{522} \\ 
 &Pred-1  & N/A  & N/A  & 0.4142  & \cellcolor{purple!30}4.72e-23   \\  
 &Pred+Corr-1 & N/A  & N/A  &\cellcolor{green!30}0.4353  &\cellcolor{purple!30}1.52e-25   \\
 \cline{2-7}
 &LDDMM-2 &N/A  &N/A &0.4409 &\cellcolor{purple!30}2.77e-26  &\multirow{3}{*}{523} \\ 
 &Pred-2  & N/A & N/A &0.4280  &\cellcolor{purple!30}1.05e-24   \\  
 &Pred+Corr-2  & N/A  & N/A &\cellcolor{green!30}0.4445 &\cellcolor{purple!30}9.64e-27  \\
 \hline
  \multirow{ 6}{*}{6mo} &LDDMM-1 &-0.4989  &\cellcolor{purple!30}8.01e-31 &0.4688 &\cellcolor{purple!30}6.09e-27 &\multirow{3}{*}{468} \\ 
 &Pred-1  &-0.4768  &\cellcolor{purple!30}6.22e-28  &0.4625  &\cellcolor{purple!30}3.47e-26   \\  
 &Pred+Corr-1 &\cellcolor{green!30}-0.5128  &\cellcolor{purple!30}9.64e-33  &\cellcolor{green!30}0.4846  &\cellcolor{purple!30}6.19e-29   \\
 \cline{2-7}
 &LDDMM-2 &\cellcolor{red!30}-0.5072  &\cellcolor{purple!30}4.29e-32 &0.4883 &\cellcolor{purple!30}1.58e-29  &\multirow{3}{*}{470} \\ 
 &Pred-2  &-0.4718  &\cellcolor{purple!30}2.02e-27 &0.4742 &\cellcolor{purple!30}9.96e-28   \\  
 &Pred+Corr-2  &-0.5066  &\cellcolor{purple!30}5.25e-32 &\cellcolor{green!30}0.4913 &\cellcolor{purple!30}6.33e-30  \\
 \hline
 \multirow{ 6}{*}{12mo} &LDDMM-1 &-0.4756  &\cellcolor{purple!30}1.43e-27 &0.4859 &\cellcolor{purple!30}7.22e-29 &\multirow{3}{*}{464} \\ 
 &Pred-1  &-0.4530  &\cellcolor{purple!30}7.32e-25  &0.4771  &\cellcolor{purple!30}9.39e-28    \\  
 &Pred+Corr-1 &\cellcolor{green!30}-0.4908  &\cellcolor{purple!30}1.67e-29  &\cellcolor{green!30}0.5064  &\cellcolor{purple!30}1.37e-31   \\
 \cline{2-7}
 &LDDMM-2 &-0.4937  &\cellcolor{purple!30}1.07e-29 &0.5026 &\cellcolor{purple!30}7.05e-31  &\multirow{3}{*}{461} \\ 
 &Pred-2  &-0.4626  &\cellcolor{purple!30}7.94e-26 &0.4913 &\cellcolor{purple!30}2.21e-29   \\  
 &Pred+Corr-2  &\cellcolor{green!30}-0.4987  &\cellcolor{purple!30}2.35e-30 &\cellcolor{green!30}0.5149 &\cellcolor{purple!30}1.44e-32  \\
 \hline
 \multirow{ 6}{*}{24mo} &LDDMM-1 &\cellcolor{red!30}-0.4120  &\cellcolor{purple!30}9.53e-15 &0.4476 &\cellcolor{purple!30}2.06e-17 &\multirow{3}{*}{325} \\ 
 &Pred-1  &-0.3670  &\cellcolor{purple!30}8.51e-12  &0.4331  &\cellcolor{purple!30}2.71e-16   \\  
 &Pred+Corr-1 &-0.4109  &\cellcolor{purple!30}1.15e-14  &\cellcolor{green!30}0.4632  &\cellcolor{purple!30}1.09e-18  \\
 \cline{2-7}
 &LDDMM-2 &\cellcolor{red!30}-0.4095  &\cellcolor{purple!30}2.09e-14 &\cellcolor{red!30}0.4375 &\cellcolor{purple!30}1.93e-16  &\multirow{3}{*}{321} \\ 
 &Pred-2  &-0.3411  &\cellcolor{purple!30}3.46e-10 &0.3940 &\cellcolor{purple!30}2.29e-13  \\  
 &Pred+Corr-2  &-0.3943  &\cellcolor{purple!30}2.20e-13 &0.4336 &\cellcolor{purple!30}3.79e-16 \\
 \hline
 \multirow{ 6}{*}{36mo} &LDDMM-1 &-0.2474  &0.55 &0.2869 &0.49 &\multirow{3}{*}{8} \\ 
 &Pred-1  &-0.2474  &0.55  &0.2869  &0.49    \\  
 &Pred+Corr-1 &-0.2474  &0.55  &0.2869  &0.49  \\
 \cline{2-7}
 &LDDMM-2 &0.0935  &0.83 &0.1695 &0.69  &\multirow{3}{*}{8} \\ 
 &Pred-2  &0.0935  &0.83 &0.1695 &0.69   \\  
 &Pred+Corr-2  &0.0935  &0.83 &0.1695 &0.69  \\
 \hline

\end{tabular}
\end{adjustbox}
\caption{FPSGR-derived correlations with clinical variables, compared to correlations with clinical variables for SGR using optimization-based LDDMM. The \#data column lists the number of data points analyzed. {\setlength{\fboxsep}{0pt}\colorbox{green!30}{\strut Green}} indicates that FPSGR using the prediction+correction network shows the strongest correlations; {\setlength{\fboxsep}{0pt}\colorbox{yellow!30}{\strut Yellow}} indicates that FPSGR using the prediction network alone shows the strongest correlations; {\setlength{\fboxsep}{0pt}\colorbox{red!30}{\strut Red}} indicates that LDDMM SGR shows the strongest correlations. The MMSE column lists correlations between atrophy scores and the mini-mental state exam scores; the DX column lists correlations between atrophy score and diagnostic category. Finally, the $p$-value column(s) list the $p$-values for the null-hypothesis that there is no correlation. Benjamini-Hochberg procedure was employed to reduce the false discovery rate and {\setlength{\fboxsep}{0pt}\colorbox{purple!30}{\strut Purple}} highlight indicates statistically significant. FPSGR using the prediction+correction network generally improves performance over using the  prediction network alone and frequently even performs slightly better than the SGR results obtained by optimization-based LDDMM.} 
\label{tab:correlation_with_clinical_variables}

\end{center}
\end{table}

\begin{table}[!t] 
\centering
\normalsize
\begin{adjustbox}{max width=0.47\textwidth}
\begin{tabular}{ |c|c|c|c|}
\hline
\multicolumn{4}{|c|}{\textbf{Normality Test}}\\
\hline 
MMSE & LDDMM & Pred & Pred+Corr\\ 
\hline 
LDDMM&  N/A& \cellcolor{green!30}0.1507& \cellcolor{green!30}0.5361\\ 
\hline
Pred&  \cellcolor{green!30}0.1507& N/A& \cellcolor{red!30}0.0183 \\
\hline
Pred+Corr&  \cellcolor{green!30}0.5361&  \cellcolor{red!30}0.0183& N/A\\
\hline
\hline
\multicolumn{4}{|c|}{\textbf{Paired $t$-test}}\\
\hline 
MMSE & LDDMM & Pred & Pred+Corr\\ 
\hline 
LDDMM&  N/A& \cellcolor{green!30}0.0005484& 0.09469173\\ 
\hline
Pred&  0.9994516& N/A& 0.9999718 \\
\hline
Pred+Corr&  \cellcolor{green!30}0.0530827&  \cellcolor{red!30}0.0000282& N/A\\
\hline 
\hline
\multicolumn{4}{|c|}{\textbf{Normality Test}}\\
\hline 
DX & LDDMM & Pred & Pred+Corr\\ 
\hline 
LDDMM&  N/A& \cellcolor{green!30}0.1963& \cellcolor{green!30}0.2356\\ 
\hline
Pred& \cellcolor{green!30}0.1963& N/A& \cellcolor{green!30}0.3208 \\
\hline
Pred+Corr& \cellcolor{green!30}0.2356& \cellcolor{green!30}0.3208& N/A\\
\hline
\hline
\multicolumn{4}{|c|}{\textbf{Paired $T$-test}}\\
\hline 
DX & LDDMM & Pred & Pred+Corr\\ 
\hline 
LDDMM&  N/A& \cellcolor{green!30}0.0010944& 0.9813582\\ 
\hline
Pred& 0.9989056& N/A& 0.9999869 \\
\hline
Pred+Corr& \cellcolor{green!30}0.0186418& \cellcolor{green!30}0.0000131& N/A\\
\hline 
\end{tabular} 
\end{adjustbox}
\caption{Results of a Shapiro-Wilk normality test and a paired $t$-test on MMSE and DX correlations among optimization-based LDDMM, FPSGR without prediction network and FPSGR with correction network. The null-hypothesis for the Shapiro-Wilk normality test is that the difference between column-method and row-method is normally distributed. The null-hypothesis for the paired $t$-test is that the column-method is statistically better than row-method. {\setlength{\fboxsep}{0pt}\colorbox{green!30}{\strut Green}} highlighted $p$-values indicate no rejection of the normality hypothesis (at 5\% significance) and thus facilitate the paired $t$-test. $p$-values highlighted in {\setlength{\fboxsep}{0pt}\colorbox{red!30}{\strut red}} indicate a rejection of the normality null-hypothesis and consequently do not allow a paired $t$-test.}
\label{tab:mmse_dx_t_test}
\end{table}

\begin{figure*}[!t]
\begin{center}
\includegraphics[width=0.8\textwidth]{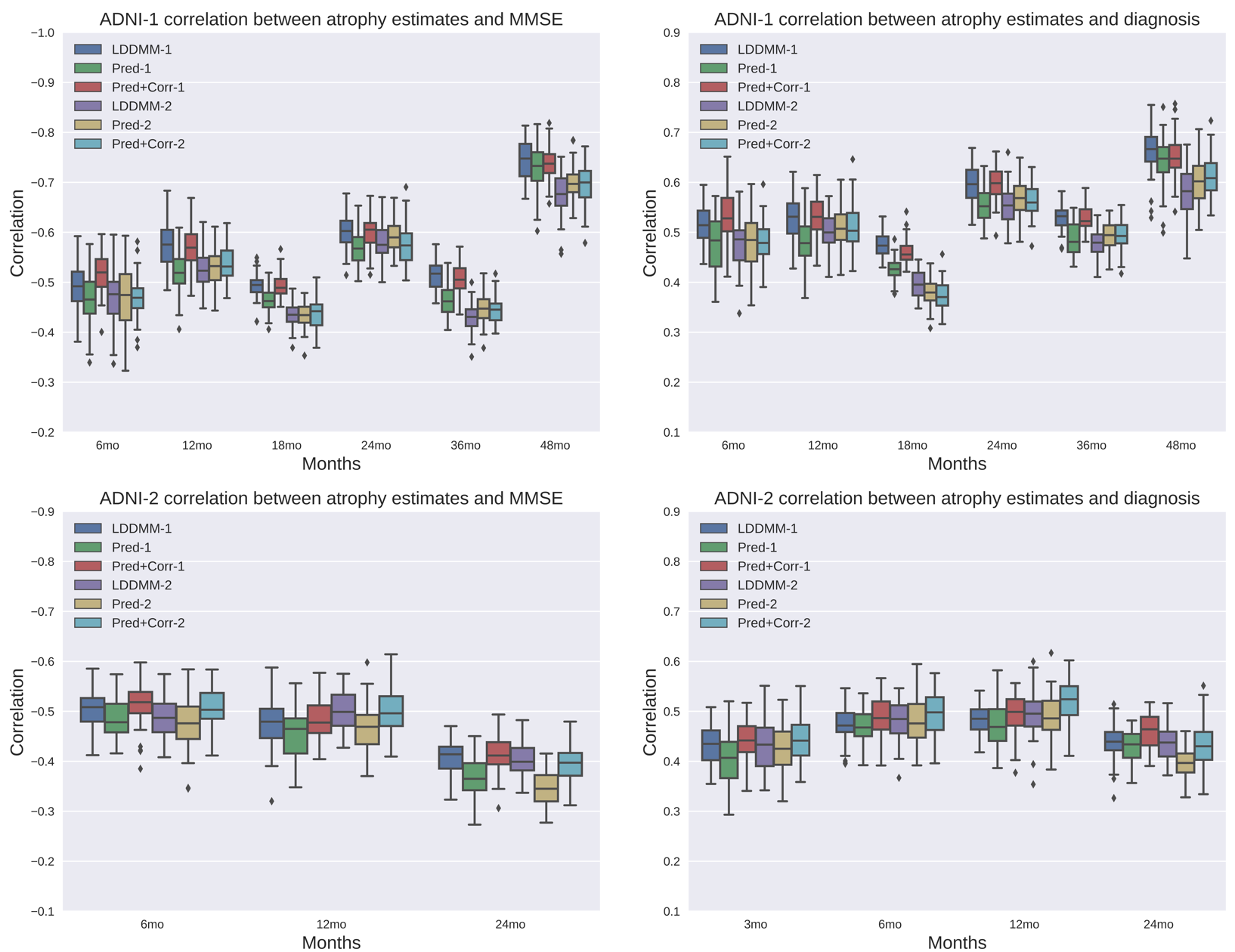}
\caption{Boxplot of FPSGR-derived correlations with clinical variables in ADNI-1 and ADNI-2. Prediction results are comparable with optimization-based LDDMM. Adding the correction network generally improves prediction results.}
\label{fig:correlations_with_clinical_variables}
\end{center}
\end{figure*}

\begin{figure*}[!phtb]
\centerline{\includegraphics[width=0.92\textwidth]{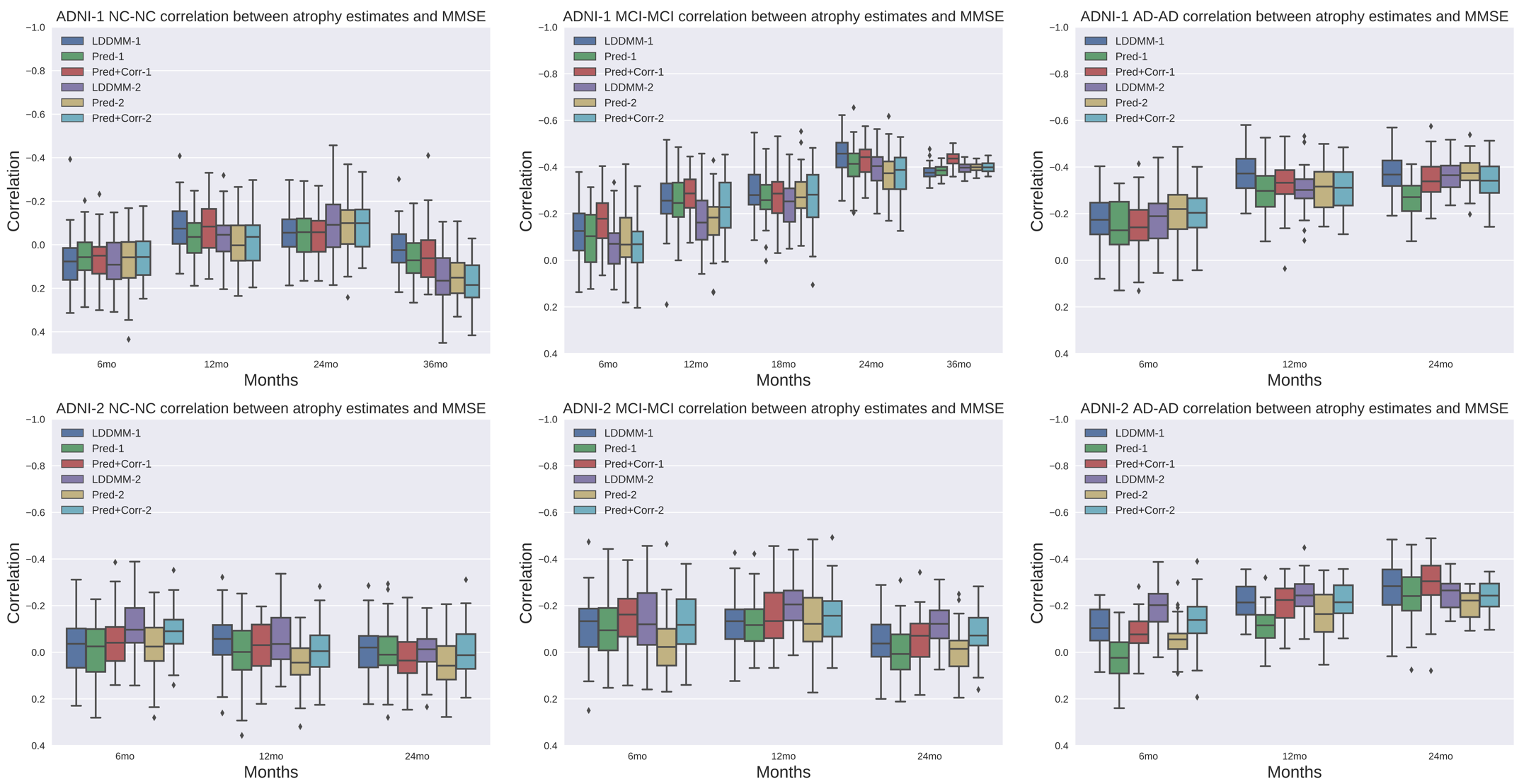}}
\caption{Boxplot of Spearman rank-order correlations between atrophy measures and MMSE with respect to time in ADNI-1 and ADNI-2. {\bf Top row:} ADNI-1 NC-NC group (left), ADNI-1 MCI-MCI group (middle), ADNI-1 AD-AD group (right). {\bf Bottom row:} ADNI-2 NC-NC group (left), ADNI-2 MCI-MCI group (middle), ADNI-2 AD-AD group (right). ADNI-1 MCI-MCI and ADNI-1 AD-AD show stronger correlations with time. In comparison, correlations remain relatively stable over time for the diagnostic groups in ADNI-2.}
\label{fig:dx}
\end{figure*}

\begin{figure*}[!phtb]
\centerline{\includegraphics[width=\textwidth]{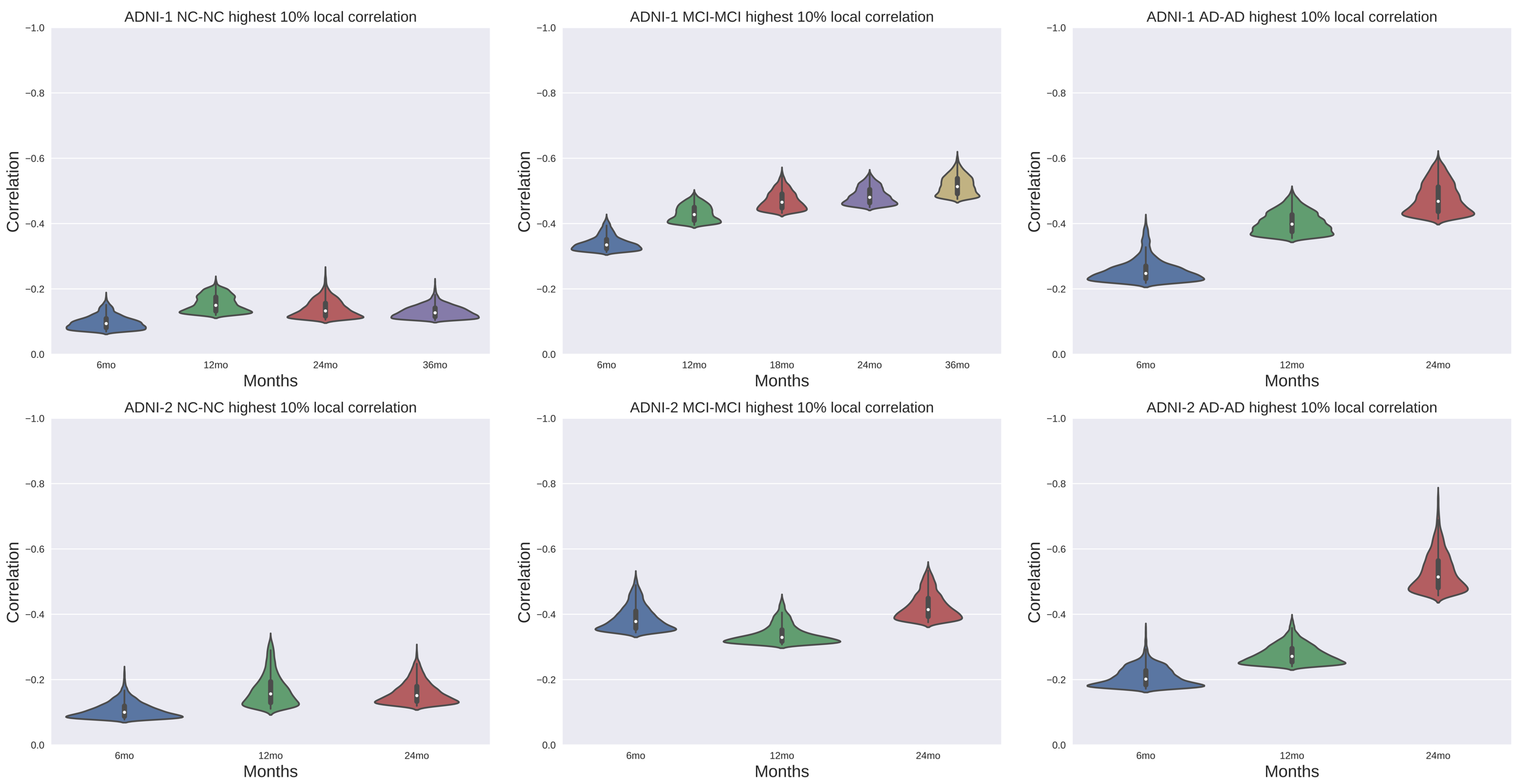}}
\caption{Kernel density estimates of highest 10\% local correlations of atrophy with MMSE within the ROI depicted in Fig.~\ref{fig:ROI}. {\bf Top row:} results of NC group, MCI group and AD group from ADNI-1. {\bf Bottom row:} results of NC group, MCI group and AD group from ADNI-2. Results show a shifting pattern for the ADNI-1 MCI case, the ADNI-1 AD case and the ADNI-2 AD case.}
\label{fig:hist}
\end{figure*}

Atrophy estimates have also been shown to correlate with clinical variables~\cite{GMF_ISBI_matching}. To quantify this effect, we computed the Spearman rank-order correlation\footnote{We used Spearman rank-order correlation instead of Pearson correlation, because the diagnostic groups imply an ordering only.} between our atrophy estimates and the diagnostic groups (NC = 0, MCI = 1, AD = 2), and also between our atrophy estimates and the scores of the mini-mental state exam (MMSE). We applied the Benjamini-Hochberg procedure~\cite{FDR} for all the correlation results in this paper to reduce the false discovery rate for multiple comparisons. The overall false discovery rate was set to be 0.01, which resulted in an effective significance level of $\alpha \approx$ 0.0093. Detailed results can be found in Table~\ref{tab:correlation_with_clinical_variables} and Fig.~\ref{fig:correlations_with_clinical_variables}, respectively. In detail, for \texttt{ADNI-1/2}, we randomly selected 200
\footnote{In \texttt{ADNI-1} 48 month, the number was 60 because there was not enough data; \texttt{ADNI-2} 36 month was omitted due to lack of data.} cases from each diagnostic category at each month and calculated the Spearman rank-order correlation. Fig.~\ref{fig:correlations_with_clinical_variables} shows the results for 50 repetitions. We observe median correlations for all four prediction models in the range of $-0.36$ to $-0.75$ for MMSE and $0.36$ to $0.65$ for diagnostic category. The correlations for all four prediction+correction models were in the range of $-0.40$ to $-0.75$ for MMSE and $0.36$ to $0.65$ for diagnostic category. Previous studies reported Pearson correlations between comparable atrophy estimates and clinical variables as high as $-0.7$ for MMSE and $0.5$ for diagnostic category for 100 subjects\cite{GMF_ISBI_matching,GMF_ISBI_optimization}.
Our two optimization-based LDDMM results achieve median correlations ranging from $-0.40$ to $-0.76$ for MMSE and $0.40$ to $0.66$ for diagnostic category, which is very similar to the predction+correlation models. In general, the correction+prediction FPSGR models outperform the models using only the prediction network. Further, using the correction network, FPSGR achieved comparable and sometimes even slightly better performance compared to the optimization-based LDDMM SGR method, see Table~\ref{tab:correlation_with_clinical_variables}
for additional quantitative results. Specifically, FPSGR using the prediction+correction network performs best in 8 out of 18 comparisons for MMSE and in 12 out of 20 comparisons for diagnostic group. In the cases where FPSGR with prediction+correction network did not perform best its difference to the best method was generally very small. In general FPSGR using the correction network performs better than FPSGR without the correction network. To check for statistical differences in the performance of FPSGR, we use a paired t-test. Table~\ref{tab:mmse_dx_t_test} shows the resulting p-values for the three methods: optimization-based SGR (i.e., LDDMM), FPSGR without correction network (i.e., Pred) and FPSGR with correction network (i.e., Pred+Corr). In both correlation with MMSE and DX, FPSGR with correction network shows significantly better performance than LDDMM and FPSGR without correction network, which justifies the use of the FPSGR method. In summary, FPSGR captures correlations between atrophy and clinical measures well.

To further explore the correlations of atrophy with MMSE scores, we visualize them separated by diagnostic group where diagnosis did not change (i.e., NC-NC, MCI-MCI, AD-AD) in Fig.~\ref{fig:dx}. For the \texttt{ADNI-1} dataset,  we observe (as expected) very low correlations for the normal diagnostic group (with no clear trend), and much stronger correlations for the MCI and AD groups. MCI and AD also exhibit increasingly stronger correlations with time. In case of \texttt{ADNI-2}, the MCI group shows modest correlations, which remain consistent across time. Correlations are relatively low for the normal groups. The AD groups show increasingly strong correlations over time. In contrast to \texttt{ADNI-1}, \texttt{ADNI-2} focuses mainly on earlier stages of the diagnostic groups \cite{Xue_ADNI2}. Hence, the deformations in \texttt{ADNI-2} are generally smaller than in \texttt{ADNI-1}. This may explain why the NC and MCI diagnostic groups show consistent correlation values over time (instead of stronger correlations as for AD in \texttt{ADNI-2} or the MCI and AD groups in \texttt{ADNI-1}).


\begin{table}[t!]
\begin{center}
\begin{adjustbox}{max width=0.47\textwidth}
\scriptsize
\begin{tabular}{c|r|c|c}
  \cellcolor{black!30}{\texttt{ADNI-1}}    &        & Slope                          & Intercept \\
 \hline
 \multirow{ 4}{*}{NC-NC} &SGR Pred-1 & [0.37, \textbf{0.44}, 0.50] & [-0.21, {\setlength{\fboxsep}{0pt}\colorbox{green!30}{\textbf{-0.08}}}, 0.05] \\ 
 &Pairwise Pred-1  & [0.44, \textbf{0.52}, 0.60] & {\setlength{\fboxsep}{0pt}\colorbox{red!30}{[-0.46, \textbf{-0.30}, -0.14]}} \\  
 \cline{2-4}
 &SGR Pred-2 & [0.43, \textbf{0.50}, 0.57] & [-0.16, {\setlength{\fboxsep}{0pt}\colorbox{green!30}{\textbf{-0.02}}}, 0.11] \\ 
 &Pairwise Pred-2  & [0.48, \textbf{0.57}, 0.65] & {\setlength{\fboxsep}{0pt}\colorbox{red!30}{[-0.34, \textbf{-0.18}, -0.01]}} \\  
 \hline
 \multirow{ 4}{*}{NC-MCI} &SGR Pred-1 & [0.39, \textbf{0.58}, 0.78] & [-0.43, {\setlength{\fboxsep}{0pt}\colorbox{green!30}{\textbf{-0.05}}}, 0.33] \\ 
 &Pairwise Pred-1  & [0.39, \textbf{0.63}, 0.87] & [-0.63, \textbf{-0.16}, 0.30] \\  
 \cline{2-4}
 &SGR Pred-2 & [0.72, \textbf{0.99}, 1.26] & [-0.68, \textbf{-0.16}, 0.36] \\ 
 &Pairwise Pred-2  & [0.65, \textbf{0.96}, 1.27] & [-0.69, {\setlength{\fboxsep}{0pt}\colorbox{yellow!30}{\textbf{-0.10}}}, 0.50] \\  
 \hline
 \multirow{ 4}{*}{MCI-MCI} &SGR Pred-1 & [0.65, \textbf{0.80}, 0.96] & [-0.29, {\setlength{\fboxsep}{0pt}\colorbox{green!30}{\textbf{-0.03}}}, 0.22] \\ 
 &Pairwise Pred-1  & [0.69, \textbf{0.86}, 1.03] & [-0.43, \textbf{-0.15}, 0.12] \\  
 \cline{2-4}
 &SGR Pred-2 & [0.69, \textbf{0.82}, 0.96] & [-0.20, {\setlength{\fboxsep}{0pt}\colorbox{green!30}{\textbf{0.02}}}, 0.24] \\
 &Pairwise Pred-2  & [0.70, \textbf{0.85}, 1.01] & [-0.29, \textbf{-0.04}, 0.21] \\  
 \hline
 \multirow{ 4}{*}{MCI-NC} &SGR Pred-1 & [0.26, \textbf{0.44}, 0.62] & [-0.61, {\setlength{\fboxsep}{0pt}\colorbox{green!30}{\textbf{-0.29}}}, 0.03] \\ 
 &Pairwise Pred-1  & [0.21, \textbf{0.45}, 0.68] & [-0.74, \textbf{-0.31}, 0.12] \\  
 \cline{2-4}
 &SGR Pred-2 & [0.40, \textbf{0.61}, 0.83] & [-0.62, \textbf{-0.24}, 0.14] \\
 &Pairwise Pred-2  & [0.29, \textbf{0.56}, 0.83] & [-0.61, {\setlength{\fboxsep}{0pt}\colorbox{yellow!30}{\textbf{-0.14}}}, 0.34] \\  
 \hline
\multirow{ 4}{*}{MCI-AD} &SGR Pred-1 & [1.28, \textbf{1.40}, 1.53] & [-0.24, {\setlength{\fboxsep}{0pt}\colorbox{green!30}{\textbf{-0.02}}}, 0.20] \\ 
 &Pairwise Pred-1  & [1.28, \textbf{1.42}, 1.56] & [-0.31, \textbf{-0.06}, 0.19] \\  
 \cline{2-4}
 &SGR Pred-2 & [1.42, \textbf{1.56}, 1.70] & [-0.11, \textbf{0.14}, 0.39] \\
 &Pairwise Pred-2  & [1.44, \textbf{1.60}, 1.75] & [-0.22, {\setlength{\fboxsep}{0pt}\colorbox{yellow!30}{\textbf{0.06}}}, 0.33] \\  
 \hline
 \multirow{ 4}{*}{AD-AD} &SGR Pred-1 & [1.23, \textbf{1.50}, 1.77] & [-0.13, \textbf{0.21}, 0.54] \\ 
 &Pairwise Pred-1  & [1.25, \textbf{1.55}, 1.85] & [-0.23, {\setlength{\fboxsep}{0pt}\colorbox{yellow!30}{\textbf{0.13}}}, 0.49] \\  
 \cline{2-4}
 &SGR Pred-2 & [1.56, \textbf{1.85}, 2.15] & [-0.13, {\setlength{\fboxsep}{0pt}\colorbox{green!30}{\textbf{0.22}}}, 0.57] \\
 &Pairwise Pred-2  & [1.53, \textbf{1.85}, 2.16] & [-0.15, \textbf{0.23}, 0.60] \\  
 \hline
 \cellcolor{black!30}{\texttt{ADNI-2}}  & & Slope & Intercept\\
 \hline
\multirow{ 4}{*}{NC-NC} &SGR Pred-1 & [0.41, \textbf{0.48}, 0.55] & [-0.03, {\setlength{\fboxsep}{0pt}\colorbox{green!30}{\textbf{0.04}}}, 0.12] \\ 
 &Pairwise Pred-1  & [0.25, \textbf{0.33}, 0.41] & {\setlength{\fboxsep}{0pt}\colorbox{red!30}{[0.15, \textbf{0.24}, 0.33]}} \\  
 \cline{2-4}
 &SGR Pred-2 & [0.47, \textbf{0.55}, 0.62] & [-0.03, {\setlength{\fboxsep}{0pt}\colorbox{green!30}{\textbf{0.05}}}, 0.13] \\ 
 &Pairwise Pred-2  & [0.26, \textbf{0.35}, 0.44] & {\setlength{\fboxsep}{0pt}\colorbox{red!30}{[0.22, \textbf{0.32}, 0.43]}} \\  
 \hline
\multirow{ 4}{*}{NC-MCI} &SGR Pred-1 & [0.53, \textbf{0.68}, 0.82] & [-0.14, {\setlength{\fboxsep}{0pt}\colorbox{green!30}{\textbf{0.01}}}, 0.16] \\ 
 &Pairwise Pred-1  & [0.37, \textbf{0.57}, 0.77] & [-0.06, \textbf{0.14}, 0.33] \\  
 \cline{2-4}
 &SGR Pred-2 & [0.58, \textbf{0.77}, 0.97] & [-0.19, {\setlength{\fboxsep}{0pt}\colorbox{green!30}{\textbf{0.01}}}, 0.22] \\ 
 &Pairwise Pred-2  & [0.42, \textbf{0.65}, 0.88] & [-0.07, \textbf{0.18}, 0.42] \\  
 \hline
\multirow{4}{*}{MCI-MCI} &SGR Pred-1 & [0.53, \textbf{0.61}, 0.68] & [-0.06, {\setlength{\fboxsep}{0pt}\colorbox{green!30}{\textbf{0.02}}}, 0.10] \\ 
 &Pairwise Pred-1  & [0.43, \textbf{0.52}, 0.61] & {\setlength{\fboxsep}{0pt}\colorbox{red!30}{[0.04, \textbf{0.14}, 0.23]}} \\  
 \cline{2-4}
 &SGR Pred-2 & [0.58, \textbf{0.66}, 0.73] & [-0.05, {\setlength{\fboxsep}{0pt}\colorbox{green!30}{\textbf{0.03}}}, 0.12] \\ 
 &Pairwise Pred-2  & [0.45, \textbf{0.54}, 0.63] & {\setlength{\fboxsep}{0pt}\colorbox{red!30}{[0.09, \textbf{0.19}, 0.29]}} \\  
 \hline
\multirow{ 4}{*}{MCI-NC} &SGR Pred-1 & [0.05, \textbf{0.29}, 0.52] & [-0.24, {\setlength{\fboxsep}{0pt}\colorbox{green!30}{\textbf{0.05}}}, 0.33] \\ 
 &Pairwise Pred-1  & [-0.10, \textbf{0.17}, 0.45] & [-0.12, \textbf{0.21}, 0.53] \\  
 \cline{2-4}
 &SGR Pred-2 & [0.24, \textbf{0.42}, 0.61] & [-0.17, {\setlength{\fboxsep}{0pt}\colorbox{green!30}{\textbf{0.05}}}, 0.28] \\ 
 &Pairwise Pred-2  & [0.03, \textbf{0.26}, 0.49] & {\setlength{\fboxsep}{0pt}\colorbox{red!30}{[0.02, \textbf{0.29}, 0.57]}} \\  
 \hline
 \multirow{ 4}{*}{MCI-AD} &SGR Pred-1 & [1.09, \textbf{1.27}, 1.46] & [-0.12, {\setlength{\fboxsep}{0pt}\colorbox{green!30}{\textbf{0.09}}}, 0.30] \\ 
 &Pairwise Pred-1  & [0.88, \textbf{1.10}, 1.32] & {\setlength{\fboxsep}{0pt}\colorbox{red!30}{[0.08, \textbf{0.33}, 0.58]}} \\  
 \cline{2-4}
 &SGR Pred-2 & [1.15, \textbf{1.35}, 1.56] & [-0.09, {\setlength{\fboxsep}{0pt}\colorbox{green!30}{\textbf{0.14}}}, 0.36] \\ 
 &Pairwise Pred-2  & [0.89, \textbf{1.13}, 1.37] & {\setlength{\fboxsep}{0pt}\colorbox{red!30}{[0.18, \textbf{0.44}, 0.70]}} \\  
 \hline
\multirow{ 4}{*}{AD-AD} &SGR Pred-1 & [1.74, \textbf{1.90}, 2.07] & [-0.09, {\setlength{\fboxsep}{0pt}\colorbox{green!30}{\textbf{0.04}}}, 0.18] \\ 
 &Pairwise Pred-1  & [1.57, \textbf{1.77}, 1.96] & {\setlength{\fboxsep}{0pt}\colorbox{red!30}{[0.01, \textbf{0.17}, 0.34]}} \\  
 \cline{2-4}
 &SGR Pred-2 & [1.97, \textbf{2.14}, 2.31] & [-0.07, {\setlength{\fboxsep}{0pt}\colorbox{green!30}{\textbf{0.07}}}, 0.21] \\ 
 &Pairwise Pred-2  & [1.79, \textbf{1.99}, 2.19] & {\setlength{\fboxsep}{0pt}\colorbox{red!30}{[0.05, \textbf{0.21}, 0.37]}} \\  
\hline

\end{tabular}
\end{adjustbox}
\caption{SGR prediction model compared with a pairwise prediction model. Slope and intercept values for simple linear regression of volume change over time. The notation for slope and intercept columns indicates [Lower bound of 95\% C.I., \textbf{point estimate}, Upper bound of 95\% C.I.]. {\setlength{\fboxsep}{0pt}\colorbox{green!30}{\strut Green}} indicates that the intercept is closer to zero (also, zero is within the 95\% confidence interval) for SGR prediction model; {\setlength{\fboxsep}{0pt}\colorbox{yellow!30}{\strut Yellow}}  indicates that the intercept is closer to zero for pairwise prediction model; {\setlength{\fboxsep}{0pt}\colorbox{red!30}{\strut Red}} indicates that the point estimate is either biased to overestimate or underestimate volume change. The SGR prediction model performs better than the pairwise prediction model.} 
\label{tab:sgr_versus_pairwise_slope_and_intercept}

\end{center}
\end{table}

\begin{table}[!t]
\begin{center}
\begin{adjustbox}{max width=0.47\textwidth}
\scriptsize
\begin{tabular}{r|r|c|c|c|c|c}
 \cellcolor{black!30}{\texttt{ADNI-1}}      &        & MMSE        &$p$-value                  & DX  &$p$-value  &\#data\\
 \hline
 \multirow{ 4}{*}{6mo} &SGR Pred-1  & \cellcolor{green!30}-0.4642 & \cellcolor{purple!30}8.09e-34 & \cellcolor{green!30}0.4754 & \cellcolor{purple!30}1.30e-35 & \multirow{2}{*}{608}\\ 
 &Pairwise Pred-1  & -0.3138 & \cellcolor{purple!30}2.31e-15 & 0.3369 & \cellcolor{purple!30}1.32e-17  \\  
 \cline{2-7}
 &SGR Pred-2  &\cellcolor{green!30}-0.4711  &\cellcolor{purple!30}8.48e-35 &\cellcolor{green!30}0.4849 &\cellcolor{purple!30}4.58e-37 &\multirow{2}{*}{606} \\ 
 &Pairwise Pred-2  &-0.3431  &\cellcolor{purple!30}3.51e-18 &0.3680 &\cellcolor{purple!30}7.24e-21  \\  
 \hline
 \multirow{ 4}{*}{12mo} &SGR Pred-1  &\cellcolor{green!30}-0.5328  &\cellcolor{purple!30}9.46e-43  &\cellcolor{green!30}0.4898  &\cellcolor{purple!30}1.97e-35 &\multirow{2}{*}{565} \\ 
 &Pairwise Pred-1  &-0.4393  &\cellcolor{purple!30}4.67e-28  &0.3996  &\cellcolor{purple!30}4.51e-23  \\  
 \cline{2-7}
 &SGR Pred-2  &\cellcolor{green!30}-0.5351  &\cellcolor{purple!30}9.79e-43 &\cellcolor{green!30}0.5120 &\cellcolor{purple!30}1.11e-38  &\multirow{2}{*}{560} \\ 
 &Pairwise Pred-2  &-0.4465  &\cellcolor{purple!30}9.61e-29 &0.4154 &\cellcolor{purple!30}1.00e-24  \\  
 \hline
 \multirow{ 4}{*}{18mo} &SGR Pred-1  &\cellcolor{green!30}-0.4659  &\cellcolor{purple!30}3.18e-14  & \cellcolor{green!30}0.4313 & \cellcolor{purple!30}3.37e-12 &\multirow{2}{*}{238} \\ 
 &Pairwise Pred-1  &-0.4164  &\cellcolor{purple!30}2.12e-11  & 0.3882  & \cellcolor{purple!30}5.56e-10  \\  
 \cline{2-7}
 &SGR Pred-2  &\cellcolor{green!30}-0.4389  &\cellcolor{purple!30}9.06e-13 & \cellcolor{green!30}0.3818 & \cellcolor{purple!30}8.80e-10 &\multirow{2}{*}{241} \\ 
 &Pairwise Pred-2  &-0.4078  &\cellcolor{purple!30}4.52e-11 & 0.3356 & \cellcolor{purple!30}9.38e-8 \\  
 \hline
 \multirow{ 4}{*}{24mo} &SGR Pred-1  &-0.5664  &\cellcolor{purple!30}2.83e-38  &0.5607  &\cellcolor{purple!30}2.18e-37 &\multirow{2}{*}{435} \\ 
 &Pairwise Pred-1  &\cellcolor{yellow!30}-0.5805  &\cellcolor{purple!30}1.51e-40  &\cellcolor{yellow!30}0.5791  &\cellcolor{purple!30}2.55e-40  \\  
 \cline{2-7}
 &SGR Pred-2  &-0.5911  &\cellcolor{purple!30}1.41e-41 &0.5714 &\cellcolor{purple!30}2.26e-38  &\multirow{2}{*}{427} \\ 
 &Pairwise Pred-2  &\cellcolor{yellow!30}-0.5927  &\cellcolor{purple!30}7.34e-42 &\cellcolor{yellow!30}0.5811 &\cellcolor{purple!30}6.26e-40  \\  
 \hline
 \multirow{ 4}{*}{36mo} &SGR Pred-1  &\cellcolor{green!30}-0.4731  &\cellcolor{purple!30}7.38e-17  &\cellcolor{green!30}0.4926  &\cellcolor{purple!30}2.42e-18 &\multirow{2}{*}{277} \\ 
 &Pairwise Pred-1  &-0.4470  &\cellcolor{purple!30}5.20e-15  &0.4798  &\cellcolor{purple!30}2.36e-17  \\  
 \cline{2-7}
 &SGR Pred-2  &-0.4425  &\cellcolor{purple!30}1.07e-13 &0.4894 &\cellcolor{purple!30}7.99e-17  &\multirow{2}{*}{256} \\ 
 &Pairwise Pred-2  &\cellcolor{yellow!30}-0.4538  &\cellcolor{purple!30}2.08e-14 &\cellcolor{yellow!30}0.4990 &\cellcolor{purple!30}1.59e-17  \\  
 \hline
 \multirow{ 4}{*}{48mo} &SGR Pred-1  &\cellcolor{green!30}-0.7294  &\cellcolor{purple!30}1.18e-12  &\cellcolor{green!30}0.6458  &\cellcolor{purple!30}2.08e-9 &\multirow{2}{*}{69} \\ 
 &Pairwise Pred-1  &-0.7100  &\cellcolor{purple!30}8.43e-12  &0.6168  &\cellcolor{purple!30}1.67e-8  \\  
 \cline{2-7}
 &SGR Pred-2  &\cellcolor{green!30}-0.6995  &\cellcolor{purple!30}9.08e-11 &\cellcolor{green!30}0.6048 &\cellcolor{purple!30}9.49e-8  & \multirow{2}{*}{65} \\ 
 &Pairwise Pred-2  &-0.6709  &\cellcolor{purple!30}9.65e-10 &0.5924 &\cellcolor{purple!30}2.01e-7  \\  
 \hline
 \cellcolor{black!30}{\texttt{ADNI-2}}       &        & MMSE        &p-value                  & DX  &p-value  &\#data\\
 \hline
 \multirow{ 4}{*}{3mo} &SGR Pred-1  & N/A  & N/A  & \cellcolor{green!30}0.4142  & \cellcolor{purple!30}4.72e-23 &\multirow{2}{*}{522} \\ 
 &Pairwise Pred-1  & N/A  & N/A  & 0.1744  & \cellcolor{purple!30}6.17e-5   \\  
 \cline{2-7}
 &SGR Pred-2  & N/A & N/A &\cellcolor{green!30}0.4280  &\cellcolor{purple!30}1.05e-24  &\multirow{2}{*}{523} \\ 
 &Pairwise Pred-2  & N/A & N/A &0.1503  &\cellcolor{purple!30}5.64e-4   \\  
 \hline
  \multirow{ 4}{*}{6mo} &SGR Pred-1  &\cellcolor{green!30}-0.4768  &\cellcolor{purple!30}6.22e-28  &\cellcolor{green!30}0.4625  &\cellcolor{purple!30}3.47e-26 &\multirow{2}{*}{468} \\ 
 &Pairwise Pred-1  &-0.3378  &\cellcolor{purple!30}5.93e-14  &0.2633  &\cellcolor{purple!30}7.29e-9   \\  
 \cline{2-7}
 &SGR Pred-2  &\cellcolor{green!30}-0.4718  &\cellcolor{purple!30}2.02e-27 &\cellcolor{green!30}0.4742 &\cellcolor{purple!30}9.96e-28  &\multirow{2}{*}{470} \\ 
 &Pairwise Pred-2  &-0.3312  &\cellcolor{purple!30}1.70e-13 &0.2849 &\cellcolor{purple!30}3.14e-10   \\  
 \hline
 \multirow{ 4}{*}{12mo} &SGR Pred-1  &\cellcolor{green!30}-0.4530  &\cellcolor{purple!30}7.32e-25  &\cellcolor{green!30}0.4771  &\cellcolor{purple!30}9.39e-28 &\multirow{2}{*}{464} \\ 
 &Pairwise Pred-1  &-0.4305  &\cellcolor{purple!30}2.34e-22  &0.4472  &\cellcolor{purple!30}3.40e-24    \\  
 \cline{2-7}
 &SGR Pred-2  &\cellcolor{green!30}-0.4626  &\cellcolor{purple!30}7.94e-26 &\cellcolor{green!30}0.4913 &\cellcolor{purple!30}2.21e-29 &\multirow{2}{*}{461} \\ 
 &Pairwise Pred-2  &-0.4223  &\cellcolor{purple!30}2.30e-21 &0.4374 &\cellcolor{purple!30}5.72e-23   \\  
 \hline
 \multirow{ 4}{*}{24mo} &SGR Pred-1  &-0.3670  &\cellcolor{purple!30}8.51e-12  &0.4331  &\cellcolor{purple!30}2.71e-16 &\multirow{2}{*}{325} \\ 
 &Pairwise Pred-1  &\cellcolor{yellow!30}-0.3772  &\cellcolor{purple!30}3.06e-12  &\cellcolor{yellow!30}0.4515  &\cellcolor{purple!30}9.99e-18   \\  
 \cline{2-7}
 &SGR Pred-2  &-0.3411  &\cellcolor{purple!30}3.46e-10 &0.3940 &\cellcolor{purple!30}2.29e-13  &\multirow{2}{*}{321} \\ 
 &Pairwise Pred-2  &\cellcolor{yellow!30}-0.3517  &\cellcolor{purple!30}8.89e-11 &\cellcolor{yellow!30}0.4239 &\cellcolor{purple!30}1.99e-15  \\  
 \hline
 \multirow{ 4}{*}{36mo} &SGR Pred-1  &-0.2474  &0.55  &0.4536  &0.26 &\multirow{2}{*}{8} \\ 
 &Pairwise Pred-1  &-0.1650  &0.70  &0.2869  &0.49    \\  
 \cline{2-7}
 &SGR Pred-2  &0.0935  &0.83 &0.1695 &0.69  &\multirow{2}{*}{8} \\ 
 &Pairwise Pred-2  &0.0935  &0.83 &0.2608 &0.53   \\  
 \hline

\end{tabular}
\end{adjustbox}
\caption{SGR prediction model compared with pairwise prediction model. Results show correlations with clinical variables. The \#data column lists the number of data points analyzed. {\setlength{\fboxsep}{0pt}\colorbox{green!30}{\strut Green}} indicates a stronger correlation for the SGR prediction method; {\setlength{\fboxsep}{0pt}\colorbox{yellow!30}{\strut Yellow}} indicates a stronger correlation for the pairwise model. The $p$-value column lists $p$-values for the null-hypothesis that there is no correlation. The Benjamini-Hochberg procedure was employed to reduce the false discovery rate (FDR). The {\setlength{\fboxsep}{0pt}\colorbox{purple!30}{\strut Purple}} highlight indicates statistically significant results after correction for multiple comparisons. In general, SGR prediction performs better than pairwise prediction demonstrating that regression stabilizes the correlation results. \texttt{ADNI-2} 36mo only has 8 data points and the $p$-value is greater than $0.1$, thus we ignore this month in our comparison.}
\label{tab:sgr_versus_pairwise_correlations}

\end{center}
\end{table}

\begin{table}[!ht] 
\centering
\normalsize
\begin{adjustbox}{max width=0.47\textwidth}
\begin{tabular}{ |c|c|c|}
\hline 
& Shapiro-Wilk normality test& Wilcoxon signed-rank test\\
\hline
MMSE& 0.01943 &0.0005226\\
\hline
DX & 0.03286 &0.0005083\\  
\hline 
\end{tabular} 
\end{adjustbox}
\caption{$p$-values for a Shapiro-Wilk normality test and Wilcoxon signed-rank test on MMSE and DX correlations between the SGR prediction model and the pairwise prediction model. The null-hypothesis for the Shapiro-Wilk normality test is that the difference of two methods is normally distributed (at a significance level of 5\%). 
The null-hypothesis for the Wilcoxon signed-rank test is that the pairwise prediction method is statistically better than the SGR prediction method (at a significance level of 5\%).\label{tab:sgr_pairwise_w_test}}
\end{table}



To address the question how stat-ROI specific measures behave over time, we explore how atrophy \emph{locally} (i.e., voxel-by-voxel) correlates with MMSE.
The local atrophy is defined as $$\left(1 - \text{det}(D \phi (x) ) \right) \times 100\enspace.$$ 
I.e., each voxel in a stat-ROI has an associated atrophy score. Fig.~\ref{fig:hist} shows kernel density estimates of the highest 10\% local correlations in a violin plot. For the 
\texttt{ADNI-1} MCI and AD groups, a clear shift toward stronger correlations can be observed over time, consistent with the boxplots of Fig.~\ref{fig:dx}. This indicates the progression of the disease. Correlations for the normal groups in 
\texttt{ADNI 1/2} are mostly centered around a modest correlation (as expected). In \texttt{ADNI-2}, only the AD diagnostic group shows a shift towards stronger correlations over time. All the other diagnostic groups show a relatively consistent distribution over time. This is also consistent with Fig.~\ref{fig:dx}.

\subsection{Justification of SGR}
\label{sec:sgr_justification}

For simple geodesic regression to be a useful model it should outperform pairwise image registration. The main conceptual difference is that the regression model will recover an \emph{average trend} based on multiple image time-points, i.e., the resulting regression geodesic will be a compromise between all the measurements. In contrast, for pairwise image registration (which can be seen as a trivial case of geodesic regression with two images only) the deformation will in general be able to match the target image well. However, just as in linear regression, this may accentuate the effects of noise. In both setups, images can be interpolated or extrapolated based on the estimated geodesic.

\vskip0.5ex
Tables~\ref{tab:sgr_versus_pairwise_slope_and_intercept} and~\ref{tab:sgr_versus_pairwise_correlations} justify the use of SGR. Specifically, Table~\ref{tab:sgr_versus_pairwise_slope_and_intercept} shows linear regression results of atrophy measures over time as obtained via SGR (i.e., using an SGR fit over all time-points followed by atrophy computations based on the deformations of the regression geodesic) compared with atrophy measures obtained by pairwise registration. For both the \texttt{ADNI-1} and the \texttt{ADNI-2} datasets, SGR outperforms the pairwise registration approach in two aspects: (1) the estimated intercept of SGR is generally closer to zero than for the pairwise method and the intercept 95\% confidence interval is narrower; (2) 11 out of 24 of the 95\% confidence intervals of the pairwise methods show bias to either overestimate or underestimate volume change, while none of the SGR results show such significant bias. Table~\ref{tab:sgr_versus_pairwise_correlations} compares the correlations between atrophy and clinical measures (MMSE and diagnostic category) of SGR and the pairwise approach. SGR performs better than the pairwise approach in 13 out of 18 cases for MMSE and in 15 out of 20 cases for the diagnostic category. Furthermore, when the pairwise method is better than SGR, the difference is much smaller compared to the differences observed for the cases where SGR is better than the pairwise method. Also note that the pairwise method shows better performance in later months compared to earlier months. This could, for example, be because the deformations are larger for later time-points and hence the registration result becomes more stable, or because SGR is also heavily influenced by the last time-point. To address the above observation, we used a Shapiro-Wilk normality test and a Wilcoxon signed-rank test. From Table~\ref{tab:sgr_pairwise_w_test} we see that we can reject the null-hypothesis of normality and 
hence, a paired $t$-test is not appropriate. As an alternative, we conducted a Wilcoxon signed-rank test to compare the SGR prediction model and the pairwise prediction model. Table~\ref{tab:sgr_pairwise_w_test}  shows that the SGR prediction model is statistically significantly better than the pairwise prediction model. Based on the above points, we conclude that SGR is more stable over time than the pairwise method and in general also results in stronger correlations.

\subsection{Forecasting}
\label{sec:forecast}

\begin{table}[!t]
\begin{center}
\begin{adjustbox}{max width=0.47\textwidth}
\scriptsize
\begin{tabular}{r|r|r|c|c|c|c|c}
 \cellcolor{black!30}{\texttt{ADNI-1}}      &      &      &MMSE  &p-value  & DX  &$p$-value  &\#data\\
 \hline
 \multirow{ 6}{*}{60mo}  &\multirow{6}{*}{Forecast} &LDDMM-1   &\cellcolor{red!30}-0.5242 &\cellcolor{purple!30}1.34e-13 & 0.5157 & \cellcolor{purple!30}3.85e-13 & \multirow{3}{*}{173}\\ 
 & &Pred-1   &-0.4727 &\cellcolor{purple!30}5.16e-11 & 0.4816 & \cellcolor{purple!30}1.98e-11\\ 
 & &Pred+Corr-1   &-0.5193 &\cellcolor{purple!30}2.48e-13 & \cellcolor{green!30}0.5240 & \cellcolor{purple!30}1.38e-13  \\  
 \cline{3-8}
 & &LDDMM-2   &-0.4501 &\cellcolor{purple!30}2.32e-10 & \cellcolor{red!30}0.4761 & \cellcolor{purple!30}1.43e-11 & \multirow{3}{*}{180}\\
 & &Pred-2   &-0.4527 &\cellcolor{purple!30}1.77e-10 &0.4620 &\cellcolor{purple!30}6.63e-11  \\ 
 & &Pred+Corr-2  &\cellcolor{green!30}-0.4582 &\cellcolor{purple!30}9.97e-11 &0.4652 &\cellcolor{purple!30}4.73e-11  \\  
\hline
 \multirow{ 6}{*}{72mo} &\multirow{6}{*}{Forecast} &LDDMM-1   &-0.4607 &\cellcolor{purple!30}1.60e-10 & 0.4507 & \cellcolor{purple!30}4.37e-10 & \multirow{3}{*}{174}\\
 & &Pred-1   &-0.4132 &\cellcolor{purple!30}1.45e-8 &0.4364  &\cellcolor{purple!30}1.75e-9 \\ 
 & &Pred+Corr-1   &\cellcolor{green!30}-0.4615 &\cellcolor{purple!30}1.47e-10 &\cellcolor{green!30}0.4667  &\cellcolor{purple!30}8.52e-11  \\  
 \cline{3-8}
 & &LDDMM-2   &-0.3662 &\cellcolor{purple!30}3.18e-7 & 0.4233 & \cellcolor{purple!30}2.15e-9 & \multirow{3}{*}{184}\\
 & &Pred-2   &-0.3793 &\cellcolor{purple!30}1.09e-7 &\cellcolor{yellow!30}0.4273 &\cellcolor{purple!30}1.46e-9 \\ 
 & &Pred+Corr-2   &\cellcolor{green!30}-0.3793 &\cellcolor{purple!30}1.09e-7 &0.4259 &\cellcolor{purple!30}1.67e-9  \\  
 \hline
 \multirow{ 6}{*}{84mo}  &\multirow{6}{*}{Forecast} &LDDMM-1   &\cellcolor{red!30}-0.3986 &\cellcolor{purple!30}1.40e-6 &0.4108 & \cellcolor{purple!30}6.17e-7 & \multirow{3}{*}{137}\\
 & &Pred-1   &-0.3495 &\cellcolor{purple!30}2.84e-5 & 0.4018  &\cellcolor{purple!30}1.13e-6  \\ 
 & &Pred+Corr-1    &-0.3946 &\cellcolor{purple!30}1.83e-6 & \cellcolor{green!30}0.4211  &\cellcolor{purple!30}2.98e-7  \\  
 \cline{3-8}
 & &LDDMM-2   &\cellcolor{red!30}-0.3293 &\cellcolor{purple!30}4.65e-5 & 0.3622 & \cellcolor{purple!30}6.53e-6 & \multirow{3}{*}{147}\\
 & &Pred-2   &-0.3199 &\cellcolor{purple!30}7.81e-5 & \cellcolor{yellow!30}0.3629 &\cellcolor{purple!30}6.25e-6 \\ 
 & &Pred+Corr-2   &-0.3187 &\cellcolor{purple!30}8.35e-5 &0.3609 &\cellcolor{purple!30}7.12e-6 \\  
 \hline

\end{tabular}
\end{adjustbox}
\caption{Correlations of forecasting results. The \#data column lists the number of data points analyzed. {\setlength{\fboxsep}{0pt}\colorbox{green!30}{\strut Green}} indicates that FPSGR using the prediction+correction network shows the strongest correlations; {\setlength{\fboxsep}{0pt}\colorbox{yellow!30}{\strut Yellow}} indicates that FPSGR using the prediction network alone shows the strongest correlations; {\setlength{\fboxsep}{0pt}\colorbox{red!30}{\strut Red}} indicates that LDDMM SGR shows the strongest correlations. The Benjamini-Hochberg procedure was employed to reduce the false discovery rate (FDR). The  {\setlength{\fboxsep}{0pt}\colorbox{purple!30}{\strut Purple}} highlight indicates statistically significant results after correction for multiple comparisons.}
\label{tab:forecast_leave_out}

\end{center}
\end{table}

\begin{table}[!t]
\begin{center}
\begin{adjustbox}{max width=0.47\textwidth}
\begin{tabular}{r|r|r|c|c|c|c|c}
 \cellcolor{black!30}{\texttt{ADNI-1}}      &     &   & MMSE        &p-value                  & DX  &p-value  &\#data\\
 \hline
 \multirow{ 14}{*}{36mo} &\multirow{3}{*}{Original} &LDDMM-1 &-0.5142  &\cellcolor{purple!30}4.29e-20 &0.5300 &\cellcolor{purple!30}1.81e-21 &\multirow{7}{*}{277} \\ 
 & &Pred-1  &-0.4731  &\cellcolor{purple!30}7.38e-17  &0.4926  &\cellcolor{purple!30}2.42e-18  \\  
 & &Pred+Corr-1 &-0.5069  &\cellcolor{purple!30}1.71e-19  &0.5296  &\cellcolor{purple!30}1.99e-21  \\
  \cline{2-7}
 &\multirow{2}{*}{Forecast} &Pred-1 &-0.4583 &\cellcolor{purple!30}1.09e-15 &0.4825 &\cellcolor{purple!30}1.93e-17\\
 & &Pred+Corr-1 & -0.4708 &\cellcolor{purple!30}1.42e-16 &0.4980 &\cellcolor{purple!30}1.21e-18 \\
  \cline{2-7}
  &\multirow{2}{*}{Replace} &Pred-1 &-0.4923 &\cellcolor{purple!30}3.43e-18 &0.5104 &\cellcolor{purple!30}1.21e-19 \\
 & &Pred+Corr-1 & -0.5097 &\cellcolor{purple!30}1.37e-19 &0.5375 &\cellcolor{purple!30}5.47e-22 \\
 \cline{2-8}
 &\multirow{3}{*}{Original} &LDDMM-2  &-0.4334  &\cellcolor{purple!30}3.79e-13 &0.4815 &\cellcolor{purple!30}2.93e-16  &\multirow{7}{*}{256} \\ 
 & &Pred-2  &-0.4425  &\cellcolor{purple!30}1.07e-13 &0.4894 &\cellcolor{purple!30}7.99e-17  \\  
 & &Pred+Corr-2  &-0.4393  &\cellcolor{purple!30}1.67e-13 &0.4863 &\cellcolor{purple!30}1.34e-16  \\
 \cline{2-7}
 &\multirow{2}{*}{Forecast} &Pred-2 &-0.4078 &\cellcolor{purple!30}1.36e-11 &0.4398 &\cellcolor{purple!30}1.95e-13 \\
 & &Pred+Corr-2 & -0.4005 &\cellcolor{purple!30}3.34e-11 &0.4301 &\cellcolor{purple!30}7.40e-13 \\
 \cline{2-7}
 &\multirow{2}{*}{Replace} &Pred-2 &-0.4202 &\cellcolor{purple!30}2.75e-12 &0.4635 &\cellcolor{purple!30}6.27e-15 \\
 & &Pred+Corr-2 & -0.4164 &\cellcolor{purple!30}4.51e-12 &0.4582 &\cellcolor{purple!30}1.38e-14 \\
 \hline
 \multirow{ 14}{*}{48mo} &\multirow{3}{*}{Original} &LDDMM-1 &-0.7456  &\cellcolor{purple!30}2.01e-13 &0.6635 &\cellcolor{purple!30}5.20e-10 &\multirow{7}{*}{69} \\ 
 & &Pred-1  &-0.7294  &\cellcolor{purple!30}1.18e-12  &0.6458  &\cellcolor{purple!30}2.08e-9  \\  
 & &Pred+Corr-1 &-0.7443  &\cellcolor{purple!30}2.30e-13  &0.6575  &\cellcolor{purple!30}8.43e-10  \\
  \cline{2-7}
 &\multirow{2}{*}{Forecast} &Pred-1 &-0.6332 &\cellcolor{purple!30}5.29e-9 &0.6165 &\cellcolor{purple!30}1.70e-8 \\
 & &Pred+Corr-1 & -0.6541 &\cellcolor{purple!30}1.10e-9 &0.6317 &\cellcolor{purple!30}5.86e-9 \\
   \cline{2-7}
 &\multirow{2}{*}{Replace} &Pred-1 &-0.6446 &\cellcolor{purple!30}2.27e-9 &0.6478 &\cellcolor{purple!30}1.78e-9 \\
 & &Pred+Corr-1 & -0.6668 &\cellcolor{purple!30}3.98e-10 &0.6800 &\cellcolor{purple!30}1.31e-10 \\
 \cline{2-8}
 &\multirow{3}{*}{Original} &LDDMM-2 &-0.6889  &\cellcolor{purple!30}2.25e-10 &0.5927 &\cellcolor{purple!30}1.98e-7  & \multirow{7}{*}{65} \\ 
 & &Pred-2  &-0.6995  &\cellcolor{purple!30}9.08e-11 &0.6048 &\cellcolor{purple!30}9.49e-8  \\  
 & &Pred+Corr-2  &-0.7005  &\cellcolor{purple!30}8.31e-11 &0.6067 &\cellcolor{purple!30}8.49e-8 \\
  \cline{2-7}
 &\multirow{2}{*}{Forecast} &Pred-2 &-0.6528 &\cellcolor{purple!30}3.79e-9 &0.5568 &\cellcolor{purple!30}1.46e-6 \\
 & &Pred+Corr-2 & -0.6403 &\cellcolor{purple!30}9.25e-9 &0.5460 &\cellcolor{purple!30}2.55e-6 \\
   \cline{2-7}
 &\multirow{2}{*}{Replace} &Pred-2 &-0.6334 &\cellcolor{purple!30}1.49e-8 &0.5970 &\cellcolor{purple!30}1.53e-7 \\
 & &Pred+Corr-2 & -0.6307 &\cellcolor{purple!30}1.79e-8 &0.5973 &\cellcolor{purple!30}1.50e-7 \\
 \hline

\end{tabular}
\end{adjustbox}
\caption{Forecast results compared with real data results. The \#data column lists the number of data points analyzed. The Benjamini-Hochberg procedure was employed to reduce the false discovery rate (FDR). {\setlength{\fboxsep}{0pt}\colorbox{purple!30}{\strut Purple}} highlight indicates statistically significant results after corrections for multiple comparisons. Forecast results are calculated by using SGR excluding 36mo and 48mo data points and then predicting 36mo and 48mo correlations. Results are compared based on the same dataset except for two invalid data points for the 36mo data.}
\label{tab:forecast_versus_real}

\end{center}
\end{table}

\begin{figure*}[!t]
\centering
\includegraphics[width=\textwidth]{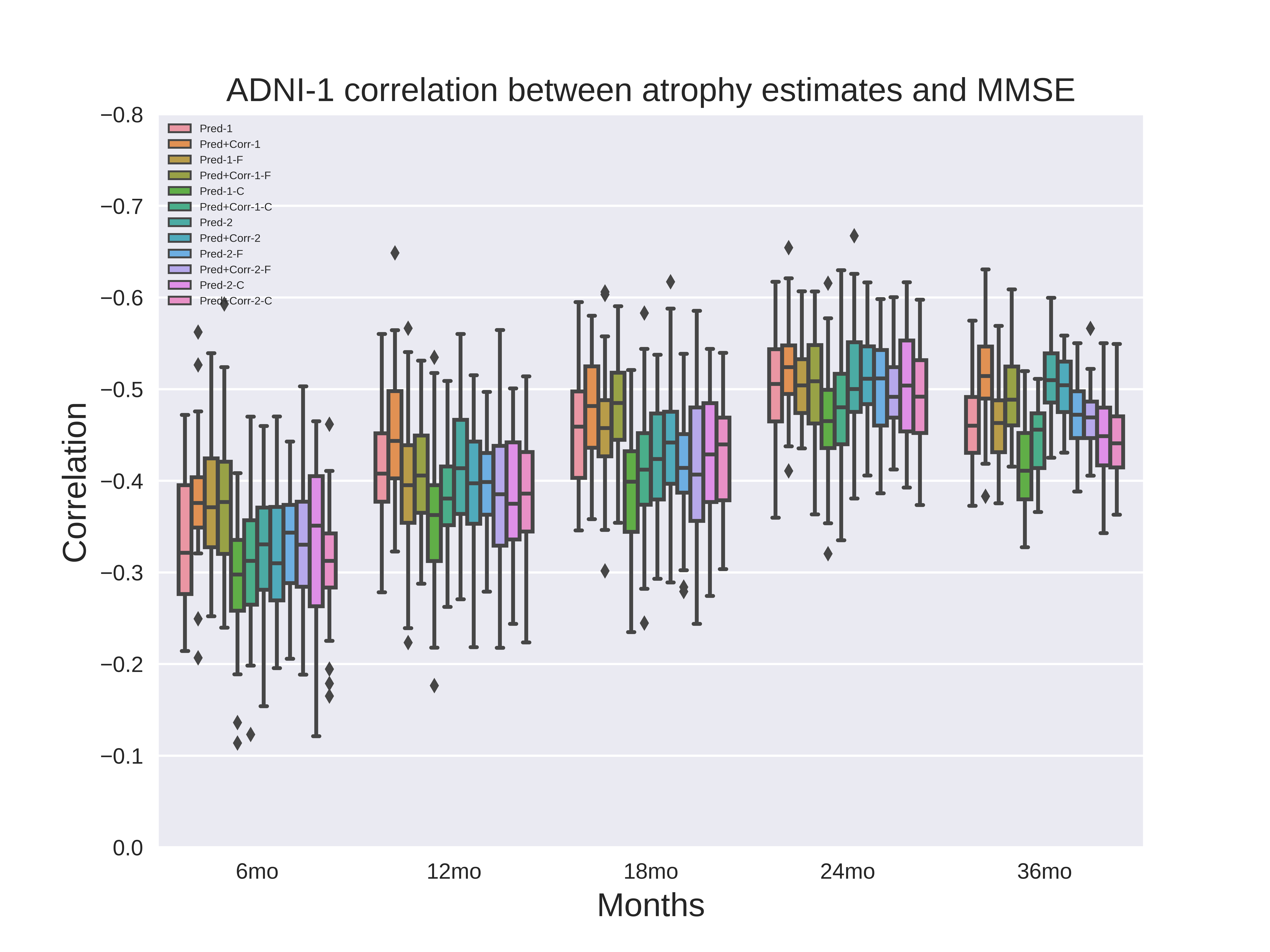}
\caption{Comparison of correlations among prediction results, \textbf{Forecast} results (-F) and \textbf{Replace} results (-C) in MCI converter groups (MCI-NC, MCI-MCI and MCI-AD). In Pred-1, Forecast results outperform Replace results; In Pred-2, Forecast results and Replace results are comparable.}
\label{fig:forecast}
\end{figure*}


Another interesting question for SGR and geodesic regression in general is the suitability of the model for the data. To address this question, we evaluate if SGR can \emph{forecast} unseen future time-points. Specifically we consider this question in two different scenarios:
\begin{itemize}
\item[\textbf{Q1})] {\bf Extrapolate-clinical:} Can we extrapolate the SGR results into the future (to time-points that do not exist in the ADNI image dataset, but for the clinical data) while still obtaining strong correlations.
\item[\textbf{Q2})] {\bf Extrapolate-image:} How well can correlations between atrophy and clinical measures be predicted for time-points when we do or do not use image data at that very time-point. We artificially leave out image measurements so that we can compare prediction results to results when we have the image measurement.
\end{itemize}

For both scenarios we use two different forecasting approaches. In the first approach ({\bf Forecast}) we simply compute SGR results with the available image time-points and then extrapolate using the resulting regression geodesic to the desired time-point in the future. In the second approach ({\bf Replace}), we artificially impute the missing image time-points by simply replacing them by the image at the closest measured time-point. For example, if we have images at 6, 12, and 18 month, but we want to forecast at 24 month, we use the 18 month image as the imputed 24 month image and then perform SGR on the 6, 12, 18, and the imputed 24 month images. We then obtain the deformation at 24 months from the SGR result.


\vskip1ex
\noindent {\bf ad Q1.} Table~\ref{tab:forecast_leave_out} shows correlations between atrophy and the clinical measures for the \textbf{Forecast} results for 60 month, 72 month and 84 month.
The resulting correlations of atrophy with diagnostic category are all above 0.3 (or below -0.3). Furthermore, the \textbf{Forecast} correlations show a downward trend with respect to time, which means that the prediction of ``far-away`` points is not as accurate as for the ``near'' future. On the other hand, SGR using the 6 month to 48 month time points results in correlation around -0.5 for MMSE and 0.5 for DX on average. Hence the correlation with the diagnostic category is consistent for that of 60 months. In other words, using 6 month to 48 month data, our prediction model can predict accurately up to 60 month. Our prediction+correction network performs as well as and even slightly better than SGR using optimization-based LDDMM. Fig.~\ref{fig:JD} shows that these forecasting results capture the trends of the changes in the temporal lobes near the hippocampus and changes in the ventricles.


\vskip1ex
\noindent {\bf ad Q2.} Table~\ref{tab:forecast_versus_real} and Fig.~\ref{fig:forecast} show \textbf{Forecast} and \textbf{Replace} results for correlations between atrophy and clinical measures in comparison to using all images. Specifically, for the \textbf{Forecast} and \textbf{Replace} results we did not use the available images at 36 and 48 month so we could compare against the results obtained when using these images. If FPSGR is a good model, it should results in correlation results as close to the correlation results using all images as possible. The \textbf{Forecast} correlations are only slightly weaker (0.02 to 0.05 lower) than the original correlations using all images illustrating that FPSGR can approximately forecast future changes. 

The overall correlations in Table~\ref{tab:forecast_versus_real} show that the \textbf{Replace} group performs better than the \textbf{Forecast} group. In particular, we are also interested in the prediction of MCI converters, namely, MCI to NC, MCI to MCI, and MCI to AD. The boxplots in Fig.~\ref{fig:forecast} show the correlations for such predictions. The \textbf{Replace} group in Fig.~\ref{fig:forecast} show relatively worse correlation performance than the \textbf{Forecast} group in ADNI-1 Pred-1 and consistent performance in ADNI-1 Pred-2. Hence SGR on a longitudinal image data can achieves good forecasting result for MCI converters.

Thus, both {\bf Extrapolate-clinical} and {\bf Extrapolate-image} experiments justify the use of FPSGR in predicting near future longitudinal trends especially for MCI converters.

\subsection{Jacobian Determinant (JD)}
\label{sec:jacobian}

\begin{figure}[!t]
\begin{center}
\includegraphics[width=0.5\textwidth]{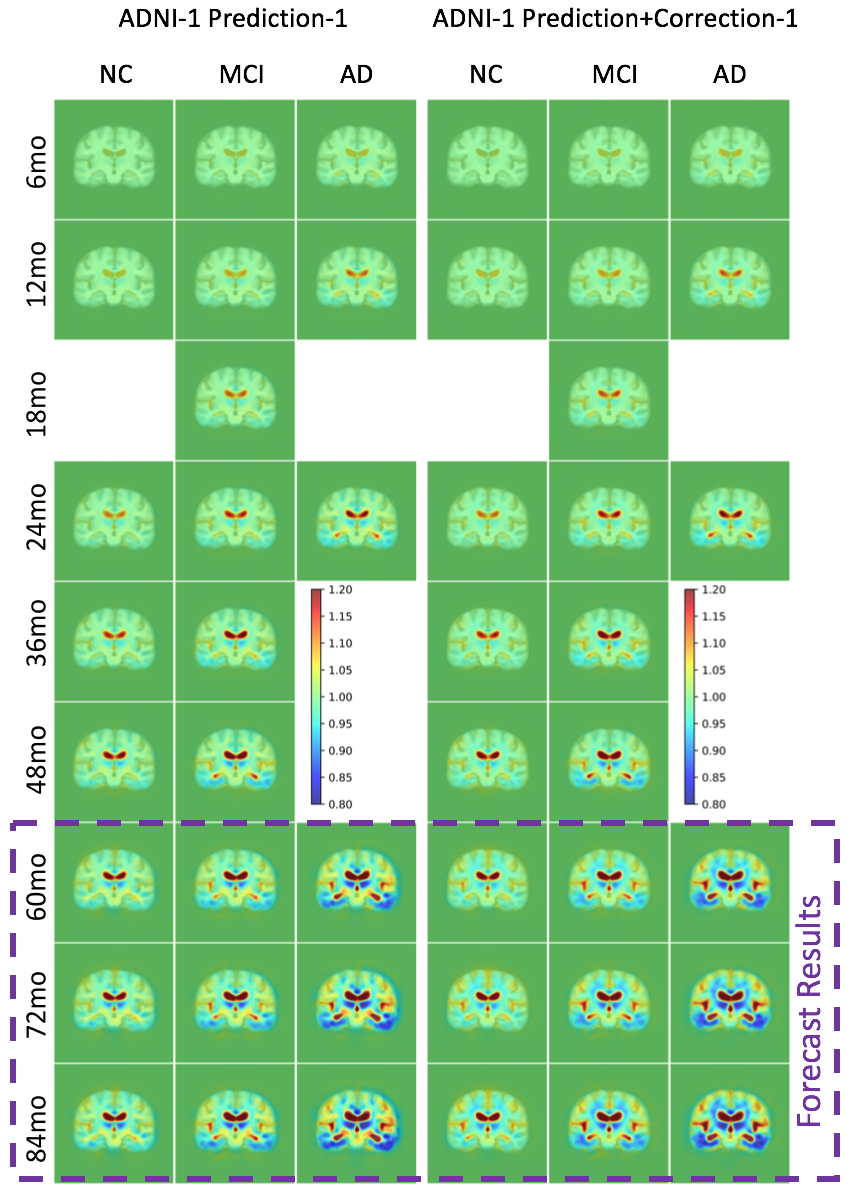}
\caption{Average Jacobian determinant over time and diagnostic category for ADNI-1 Prediction-1 and ADNI-1 Prediction+Correction-1 (experiments in ADNI-2 show similar results). A value $<$ 1 means shrinkage and value $>1$ means expansion. The 60 month - 84  month results contained in the purple rectangle are forecasts using the data from 6 month - 48 month. Results show consistent volume loss over time near the temporal lobes and expansion over time near the ventricles/cerebrospinal fluid.}
\label{fig:JD}
\end{center}
\end{figure}

The average JD images qualitatively agree with prior results~\cite{Xue_ADNI1,Xue_ADNI2}: severity of volume change increases with severity of diagnosis and time. Change is most substantial in the temporal lobes near the hippocampus (see Fig.~\ref{fig:JD}). In Fig.~\ref{fig:JD}, 6 month to 48 month are existing data points; 60 month to 84 month are forecast results. Blue indicates volume loss. Red indicates expansion. Results are consistent with expectations: volume loss increases with time and severity of diagnosis in temporal lobes; volume expansion increases with respect to time and severity of diagnosis around the ventricles / cerebrospinal fluid. The forecast results capture visually sensible volume loss or expansion over time, qualitatively illustrating the performance of our method.

\section{Conclusion and future work}
\label{sec:conclusion}
\noindent
In this work, we proposed a fast approach for geodesic regression (FPSGR) to study longitudinal image data. FPSGR incorporates the recently proposed FPIR~\cite{ref:yang2016,quicksilver} into the SGR~\cite{ref:hong} framework, thus leading to a computationally efficient solution to geodesic regression. Since FPSGR replaces the computationally intensive intermediate step of computing pairwise initial momenta via a deep-learning prediction method, it is orders of magnitude faster than existing approaches~\cite{ref:hong,hong2017fast}, without compromising accuracy. 
Consequently, FPSGR facilitates the analysis of large-scale imaging studies. Experiments on the \texttt{ADNI-1/ADNI-2} datasets demonstrate that FPSGR captures expected atrophy trends of normal aging, MCI and AD. It further (1) exhibits negligible bias towards volume changes within stat-ROIs, (2) shows high correlations with clinical variables (MMSE and diagnosis) and (3) produces consistent forecasting results on unseen data. 

In future work, it will be be interesting to explore FPSGR for the task of \emph{classifying} 
stable Mild Cognitive Impairment (sMCI) and progressive Mild Cognitive Impairment (pMCI). Currently, FPSGR only shows 
modest accuracy for distinguishing these types within MCI. Extending our approach to time-warped 
geodesic regression models~\cite{hong2014time} might improve the accuracy in this context. Furthermore,  
end-to-end prediction of averaged initial momenta would be an interesting future direction, as this 
would allow \emph{learning} representations that characterize the geodesic path among multiple time-series images, not only based on averages of momenta for two images as in FPIR~\cite{ref:yang2016,quicksilver}. 

\section*{Acknowledgements}

Research reported in this publication was supported by the National Institutes of Health (NIH) and the National Science Foundation (NSF) under award numbers NIH R01AR072013, NSF ECCS-1148870, and EECS-1711776. The content is solely the responsibility of the authors and does not necessarily represent the official views of the NIH or the NSF. We also thank Nvidia for the donation of a TitanX GPU. 

Data collection and sharing for this project was funded by the Alzheimer's Disease Neuroimaging Initiative (ADNI) (National Institutes of Health Grant U01 AG024904) and DOD ADNI (Department of Defense award number W81XWH-12-2-0012). ADNI is funded by the National Institute on Aging, the National Institute of Biomedical Imaging and Bioengineering, and through generous contributions from the following: AbbVie, Alzheimer's Association; Alzheimer's Drug Discovery Foundation; Araclon Biotech; BioClinica, Inc.; Biogen; Bristol-Myers Squibb Company; CereSpir, Inc.; Cogstate; Eisai Inc.; Elan Pharmaceuticals, Inc.; Eli Lilly and Company; EuroImmun; F. Hoffmann-La Roche Ltd and its affiliated company Genentech, Inc.; Fujirebio; GE Healthcare; IXICO Ltd.; Janssen Alzheimer Immunotherapy Research \& Development, LLC.; Johnson \& Johnson Pharmaceutical Research \& Development LLC.; Lumosity; Lundbeck; Merck \& Co., Inc.; Meso Scale Diagnostics, LLC.; NeuroRx Research; Neurotrack Technologies; Novartis Pharmaceuticals Corporation; Pfizer Inc.; Piramal Imaging; Servier; Takeda Pharmaceutical Company; and Transition Therapeutics. The Canadian Institutes of Health Research is providing funds to support ADNI clinical sites in Canada. Private sector contributions are facilitated by the Foundation for the National Institutes of Health (\url{www.fnih.org}). The grantee organization is the Northern California Institute for Research and Education, and the study is coordinated by the Alzheimer's Therapeutic Research Institute at the University of Southern California. ADNI data are disseminated by the Laboratory for Neuro Imaging at the University of Southern California.

\section*{References}

\bibliography{allBibShort}

\end{document}